\documentclass[letterpaper]{article} 
\usepackage{aaai2026}  
\usepackage{times}  
\usepackage{helvet}  
\usepackage{courier}  
\usepackage[hyphens]{url}  
\usepackage{graphicx} 
\usepackage{natbib}  
\usepackage{caption} 
\usepackage{algorithm}
\usepackage{algorithmic}
\usepackage{xspace}
\usepackage{marvosym}
\usepackage{amsmath}
\usepackage{amssymb}
\usepackage{booktabs}
\usepackage{multirow}
\usepackage{subcaption}
\newcommand{\Camera}{\textsc{Camera}\xspace}
\newcommand{\CameraP}{\textsc{Camera-P}\xspace}
\newcommand{\CameraQ}{\textsc{Camera-Q}\xspace}
\newcommand{\CameraQd}{\textsc{Camera-Q}$^\dagger$\xspace}

\usepackage{newfloat}
\usepackage{listings}
\DeclareCaptionStyle{ruled}{labelfont=normalfont,labelsep=colon,strut=off} 
\floatstyle{ruled}
\newfloat{listing}{tb}{lst}{}
\floatname{listing}{Listing}
\title{\Camera: Multi-Matrix Joint \underline{C}ompression for MoE Models vi\underline{a} \\ \underline{M}icro-\underline{E}xpert \underline{R}edundancy \underline{A}nalysis}
\author{
    Yuzhuang Xu$^{1}$,\quad Xu Han$^{2}$,\quad Yuanchi Zhang$^{3}$,\quad Yixuan Wang$^{1}$,\\
    Yijun Liu$^{1}$,\quad Shiyu Ji$^{1}$,\quad Qingfu Zhu$^{1}$,\quad Wanxiang Che$^{1,}$\textsuperscript{\Letter}
}
\affiliations{
    \textsuperscript{\rm 1}Research Center for Social Computing and Interactive Robotics, Harbin Institute of Technology, Harbin, China\\
    \textsuperscript{\rm 2}Department of Computer Science and Technology, Tsinghua University, Beijing, China\\
    \textsuperscript{\rm 3}WeChat AI, Tencent Inc, Beijing, China\\
    \{xyz, car\}@ir.hit.edu.cn
}

\usepackage{bibentry}

\begin{document}

\maketitle

\begin{abstract}

Large Language Models (LLMs) with Mixture-of-Experts (MoE) architectures are distinguished by their strong performance scaling with increasing parameters across a wide range of tasks, yet they also suffer from substantial computational and storage overheads. Notably, the performance gains of MoE models do not scale proportionally with the growth in expert parameters. While prior works attempt to reduce parameters via expert-level pruning, merging, or decomposition, they still suffer from challenges in both performance and computational efficiency. In this paper, we address these challenges by introducing \textbf{micro-expert} as a finer-grained compression unit that spans across matrices. We first establish a more fundamental perspective, viewing MoE layers as mixtures of micro-experts, and present \Camera, a lightweight and training-free framework for identifying micro-expert redundancy. Our analysis uncovers significant variance in micro-expert contributions during decoding. Based on this insight, we further propose \CameraP, a structured micro-expert pruning framework, and \CameraQ, a mixed-precision quantization idea designed for micro-experts. Extensive experiments on nine downstream tasks show that \CameraP consistently outperforms strong baselines under pruning ratios ranging from 20\% to 60\%. Furthermore, \CameraQ achieves superior results under aggressive 2-bit quantization, surpassing existing matrix- and channel-level ideas. Notably, our method enables complete micro-expert analysis of Qwen2-57B-A14B in less than 5 minutes on a single NVIDIA A100-40GB GPU.
Code is available at \url{https://github.com/xuyuzhuang11/CAMERA}.
\end{abstract}


\section{Introduction}
\label{sec:intro}

Large Language Models (LLMs) based on Mixture-of-Experts (MoE) architecture leverage sparse Feed-Forward Network (FFN) structures to enable efficient model scaling, where each time only a subset of experts is activated by a router~\citep{moe2017, switch2022}. This design facilitates the scaling of LLMs to hundreds of billions of parameters. Many well-known open-source models, such as Qwen3-MoE~\citep{qwen2025}, Deepseek-MoE~\citep{deepseek2024}, Kimi-K2~\citep{kimi2025}, and Mixtral-MoE~\citep{mixtral2024}, adopt the MoE architecture and achieve impressive results on downstream tasks. However, such substantial expansion in parameters does not bring proportional improvements in model capability, while also coupling with prohibitive computational and storage overheads—highlighting the structural redundancy inherent in MoE designs.

\begin{figure}[t]
\centering
\includegraphics[width=0.98\columnwidth]{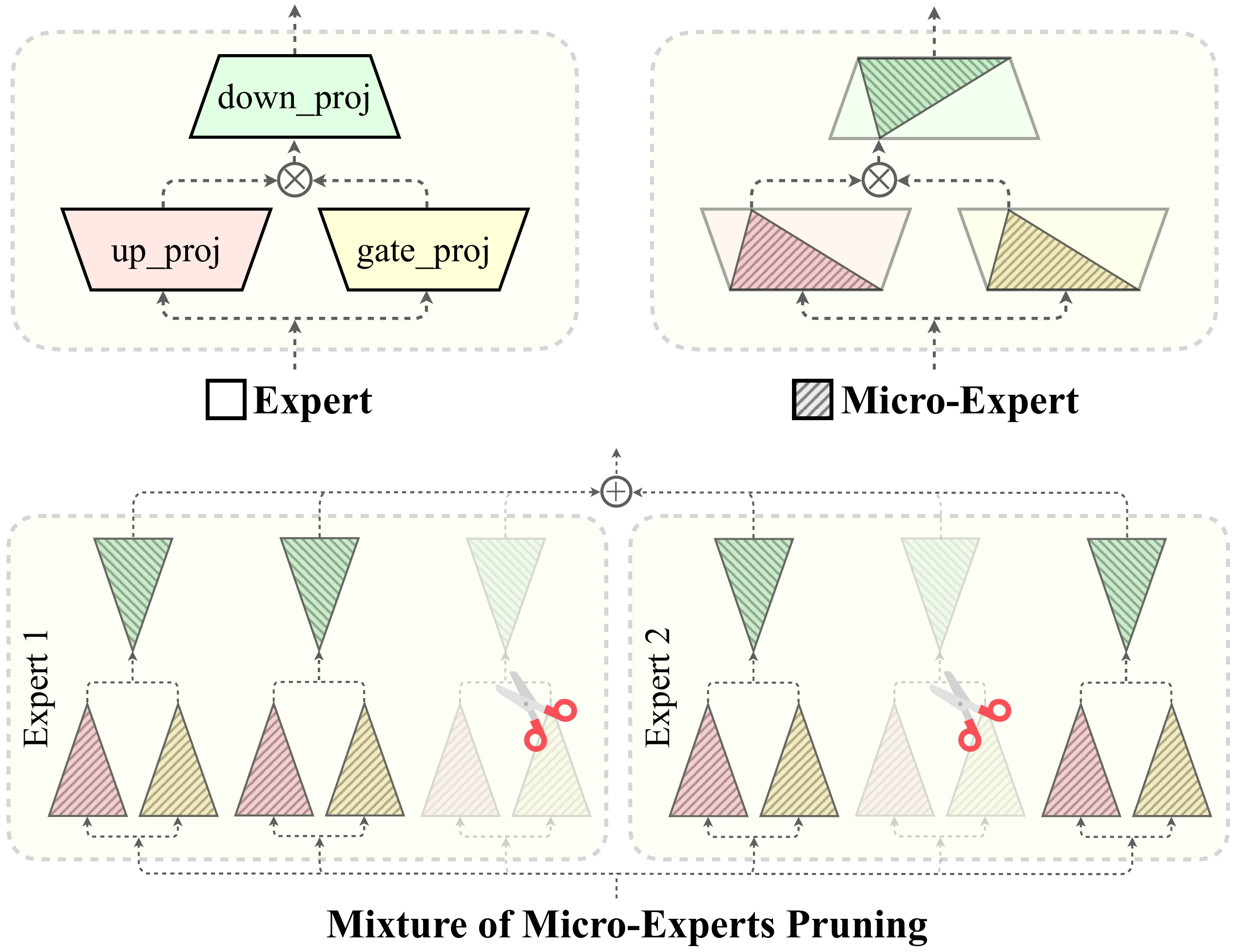}
\caption{Transition from Experts to Micro-Experts. The lower part illustrates the structure of the mixture of micro-experts and the corresponding pruning strategy.}
\label{fig:intro}
\vspace{-2em}
\end{figure}

Pruning is a widely used strategy to reduce redundancy in MoE models. Existing methods for reducing expert parameters primarily fall into two categories: expert pruning and expert merging. Expert pruning removes either all or part of the parameters within each expert~\citep{naee2024, moei2024}. Full pruning inevitably results in severe information loss, while partial pruning often yields suboptimal performance due to the lack of reliable measures for intra-expert importance. Expert merging, on the other hand, seeks to mitigate information loss by assuming functional similarity among experts~\citep{moei2024, mcsmoe2023}. However, this assumption rarely holds in practice, thereby limiting the effectiveness of merging-based strategies. 
An emerging direction exemplified by $D^2$-MoE attempts to relax this assumption by combining expert merging with delta compression. Specifically, it constructs a shared base weight and expert-specific delta weights, both compressed using low-rank approximation~\citep{delta2025,resmoe2025}.

Although recent studies suggest that partial expert pruning is more effective than full expert removal~\citep{pruner2024}, especially in modern MoE models with a large number of experts, challenges remain due to limited understanding of expert behavior within MoE layers. In particular, it is difficult to identify essential parameters and determine optimal retention ratios for each expert. 
A further limitation of most existing methods is that they compress each matrix in the expert independently, overlooking the functional dependencies among different matrices. To shed light on the internal mechanism of MoE layers, we introduce a finer-grained structural perspective by viewing each MoE layer as a mixture of micro-experts, where each micro-expert is jointly defined by the three transformations—\texttt{up\_proj}, \texttt{gate\_proj} and \texttt{down\_proj}, as shown in Figure~\ref{fig:intro}. Our analysis reveals that the output of an MoE layer is a linear combination of all micro-experts, whose relative importance varies significantly. This insight forms the critical basis for the compression strategies proposed in this work.

We frame micro-expert pruning as the problem of removing a fixed proportion (e.g., 20\%) of micro-experts while minimizing degradation in decoding performance. As we detail later, this is an NP-hard problem that cannot be solved exactly within practical time constraints. To address this challenge, we propose \Camera, an efficient and accurate approximation algorithm for estimating the importance ranking of micro-experts in MoE layers. Building on this analysis, we propose \CameraP, a structured pruning framework that jointly prunes redundant micro-experts across the three FFN weight matrices, as shown in Figure~\ref{fig:intro}. We further extend this idea to mixed-precision quantization by proposing \CameraQ, a novel micro-expert-aware partitioning scheme for assigning precision. Experiments show that \CameraP can prune MoE models ($>$50B) within 5 minutes on a single NVIDIA A100-40GB GPU—orders of magnitude faster than existing methods that often require hours of computation on multi-GPUs. Both \CameraP and \CameraQ consistently outperform strong baselines across nine zero-shot tasks and diverse MoE models, demonstrating superior efficiency and generalization. 

Overall, we summarize our key contributions as follows:
\begin{itemize}
    \item We propose \Camera, a training-free and effective approximation algorithm that accurately ranks micro-experts by their importance in MoE layers, providing the foundation for micro-expert-based compression.
    \item We propose \CameraP, a structured pruning framework that jointly prunes across the matrices in each FFN, preserving functional integrity and coordination.
    \item We present \CameraQ, a novel micro-expert–aware mixed-precision quantization idea that allocates bit-widths based on micro-expert importance.
    \item Extensive experiments on mainstream MoE models and benchmarks demonstrate that both \CameraP and \CameraQ consistently outperform strong baselines, while achieving high scalability and efficiency.
\end{itemize}

\section{Related Work}

\subsection{Model Compression}

This work focuses on model weight compression. The primary approaches for compressing LLM weights include quantization, pruning, knowledge distillation, and low-rank approximation. Most quantization methods target post-training conversion of well-trained model weights into low-bit representations~\citep{gptq2022, smooth2023}. The state-of-the-art quantization algorithms can reduce weights to 3-bit precision with negligible performance degradation~\citep{spinquant2025, omniquant2023}. With the aid of carefully designed mixed-precision strategies, even sub-3-bit quantization can maintain performance comparable to the original model~\citep{quip2024, crvq2025}. Pruning, on the other hand, reduces model size by eliminating less critical weights. Unstructured pruning offers maximum flexibility in weight removal and typically results in the least performance drop~\citep{wanda2024}. Structured pruning imposes constraints on which weights can be pruned based on their location or structure, which may lead to slightly larger performance loss but enable actual inference speedup and make it more suitable for deployment~\citep{sparsegpt2023}. Knowledge distillation guided by loss minimization~\citep{onebit2024}, and low-rank approximation typically via singular value decomposition~\citep{delta2024}, are also commonly integrated into broader compression frameworks. For further reading, please refer to survey~\cite{survey2024}. As MoE becomes a dominant architecture for scaling LLMs, several compression techniques are developed specifically for MoE-based models.

\subsection{Pruning Methods for MoE Models}

Most pruning methods for MoE models focus on expert-level compression rather than on individual matrices. These fall mainly into two categories: direct pruning and expert merging (sometimes followed by pruning). Direct pruning methods like NAEE exhaustively evaluate expert combinations and retain the one yielding the lowest loss on calibration data~\citep{naee2024}. While effective on Mixtral-MoE, this brute-force approach does not scale well to modern sparse MoEs. MoE-$I^2$~\citep{moei2024} and MoE-Pruner~\citep{pruner2024} partially prune expert weights, but struggle to balance identifying important weights and achieving speedup. Expert merging methods fall into two subgroups. The first assumes functional similarity across experts, grouping and merging them based on activation frequency, output similarity, or knowledge distribution (e.g., MC-SMoE~\citep{mcsmoe2023}, HC-SMoE~\citep{hcsmoe2024}, TAP~\cite{tap2024}, EEP~\citep{eep2024}). This idealized assumption often limits performance. The second extracts common components from all experts (e.g., $D^2$-MoE~\citep{delta2025}, ResMoE~\cite{resmoe2025}, Sub-MoE~\citep{submoe2025}, MoE-SVD~\citep{moesvd2025}), then applies low-rank compression to approximate residuals. However, these approaches overlook the functional integrity across the three transformations in FFNs and often fail to preserve the most critical parameters. Our method, \CameraP, belongs to the partial pruning family but differs in two key ways. First, it performs fine-grained pruning at the micro-expert level, capturing cross-matrix coordination in FFNs. Second, it is powered by \Camera, our fast and accurate importance estimator, addressing a major bottleneck of existing methods.

\subsection{Quantization Methods for MoE Models}

Most quantization work targets LLMs rather than MoEs, with MoE studies focusing mainly on assigning different bit-widths across experts. 
Studies like MC~\citep{mc2025}, MxMoE~\citep{mxmoe2025}, and AFGQ~\citep{afgq2025} assign bit-widths based on a combination of activation frequency and weight sensitivity. In contrast, our \CameraQ is built on \Camera, requiring no activation statistics, no pre-quantization, and no expensive evolutionary search.

\section{Micro-Expert Redundancy Analysis}
\label{sec:camera}

In this section, we first formalize the definition of micro-experts, followed by a general mathematical model that characterizes redundancy in MoE layers. We then highlight the challenges in analyzing such redundancy and derive the \Camera algorithm as an efficient approximation, with provable and controllable error bounds.

\subsection{From Expert to Micro-Expert}
\label{subsec:me}

In the standard MoE architecture, each decoder layer comprises a self-attention layer and an MoE layer. The \textit{i}-th expert in the MoE layer is typically defined as:
\begin{equation}
    E_i\left(\mathbf{x}\right) = \mathbf{W}_i^{\mathrm{down}}\left[ \sigma\left(\mathbf{W}_i^{\mathrm{gate}}\mathbf{x}\right)\cdot\mathbf{W}_i^{\mathrm{up}}\mathbf{x} \right],
\label{eq:E}
\end{equation}
where $\mathbf{x}\in \mathbb{R}^{d_\mathrm{model}}$ and $E_i(\mathbf{x})\in \mathbb{R}^{d_\mathrm{model}}$ are the input/output hidden states, $\mathbf{W}^{\mathrm{up/gate/down}}$ are the transformation matrices, and $\sigma(\cdot)$ is the SiLU activation. The MoE layer produces a weighted sum of $N_E$ expert outputs, with weights $A_i(\mathbf{x})$ determined by the router (omit Top-\textit{k}). This process is:
\begin{equation}
    \mathbf{y} = \sum_i^{N_E} A_i\left(\mathbf{x}\right)\cdot E_i\left(\mathbf{x}\right).
\label{eq:expert}
\end{equation}
We now adopt a microscopic perspective by decomposing each expert into micro-experts. Consider the weight matrices of a given expert. Let the \textit{i}-th row of $\mathbf{W}^\mathrm{up}$ and $\mathbf{W}^\mathrm{gate}$ be denoted as $\mathbf{w}_i^\mathrm{up/gate}$, and the \textit{i}-th column of $\mathbf{W}^\mathrm{down}$ as $\mathbf{w}_i^\mathrm{down}$. We then define the \textit{i}-th micro-expert as follows:
\begin{equation}
    e_{i}\left(\mathbf{x}\right) = \mathbf{w}_{i}^{\mathrm{down}}\left[ \sigma\left(\mathbf{w}_i^{\mathrm{gate}}\mathbf{x}\right)\cdot\mathbf{w}_i^{\mathrm{up}}\mathbf{x} \right].
\end{equation}
The output of the MoE layer $\mathbf{y} = \sum_i^{N_e} A_i\left(\mathbf{x}\right)\cdot e_i\left(\mathbf{x}\right)$ can be viewed as a weighted combination of all $N_e$ micro-experts, where the number of micro-experts $N_e = N_E \times d_\mathrm{ff}$, and $d_\mathrm{ff}$ is the intermediate dimension of each expert. Note that $A_i\left(\mathbf{x}\right)$, $\sigma\left(\mathbf{w}_i^{\mathrm{gate}}\mathbf{x}\right)$, and $\mathbf{w}_i^{\mathrm{up}}\mathbf{x}$ are all scalars, so we denote
\begin{equation}
    \phi_i = a_i\left(\mathbf{x}\right)\cdot\sigma\left(\mathbf{w}_i^{\mathrm{gate}}\mathbf{x}\right)\cdot\mathbf{w}_i^{\mathrm{up}}\mathbf{x}.
\end{equation}
Based on this, Equation~\ref{eq:expert} can be rewritten as:
\begin{equation}
    \mathbf{y} = \sum_i^{N_e} \phi_i\mathbf{w}_i^{\mathrm{down}}.
\label{eq:mome}
\end{equation}
For simplicity, we denote all $\mathbf{w}_i^{\mathrm{down}}$ as $\mathbf{w}_i$ in the following paper. Clearly, the function of each micro-expert comprises two parts, represented by $\phi_i$ and $\mathbf{w}_i$ respectively. The $\{\mathbf{w}_i\}_{i=1}^{n_e}$ are fixed weight vectors, which we refer to as the \textbf{basis vector set}. This indicates that the output $\mathbf{y}$ of the MoE layer is actually a linear combination of the basis vectors. For different hidden states $\mathbf{x}$, the \textbf{combination coefficients} $\phi_i$ vary and are determined by $\mathbf{x}$.

\subsection{Redundancy Problem}
\label{subsec:problem}

Let $(\mathbf{x}, \mathbf{y})$ denote the input-output hidden state tuple of the MoE layer. Given any input $\mathbf{x}$, we aim to find a subset of micro-experts such that the decoding result $\hat{\mathbf{y}}$ based on this subset deviates minimally from the original $\mathbf{y}$. The micro-experts excluded in this process are considered to exhibit the highest redundancy. We introduce a calibration dataset $\{(\mathbf{x}_i, \mathbf{y}_i)\}_{i=1}^n$ as a proxy for the ideal hidden state space. For the \textit{i}-th sample, we have:
\begin{equation}
    \mathbf{y}_i=\phi_{i1}\mathbf{w}_1+\phi_{i2}\mathbf{w}_2+\cdots+\phi_{iN_e}\mathbf{w}_{N_e}.
\end{equation}
This decoding process can be compactly expressed using matrix notation as $\mathbf{Y}=\mathbf{\Phi W}$, where $\mathbf{\Phi}\in \mathbb{R}^{n\times N_e}$ is the activation coefficient matrix of all inputs $\mathbf{x}_i$ over all micro-experts, and $\mathbf{W}\in \mathbb{R}^{N_e\times d_\mathrm{model}}$ is the matrix of basis vectors (i.e., the transpose of $\mathbf{W}^\mathrm{down}$).

Now, to decode using a selected subset of $m$ micro-experts, we aim to minimize the error between the resulting output and the original output $\mathbf{Y}$. Let $S$ denote the index set of the selected micro-experts to be retained. This optimization problem can be formally written as:
\begin{equation}
    \min_{S\subset[N_e],|S|=m}\|\mathbf{Y}-\mathbf{\Phi}_{:,S}\mathbf{W}_{S,:}\|_F^2.
\label{eq:cssp}
\end{equation}
This is similar to a class of problems known as Column Subset Selection Problems (CSSP). Such combinatorial optimization problems have been proven to be NP-hard, making it impossible to obtain exact solutions in polynomial time~\citep{cssp2021}. Although general approximate solutions exist~\citep{cur2009, iterfs2018}, the large $N_e$ still poses significant challenges for efficient hardware utilization. For instance, models like Mixtral-8$\times$7B and Deepseek-MoE-16B have $N_e$ on the order of $10^5$. Even with approximation, the time and space costs for these models remain substantial.

\subsection{\Camera Algorithm}
\label{subsec:camera}

To approximately solve the optimization problem formulated in Equation~\ref{eq:cssp}, we first consider the impact of pruning a small subset of micro-experts on the output $\mathbf{Y}$. The following lemma provides an initial result.

\paragraph{Lemma} Let $S^C$ be the index set of removed micro-experts. Then the upper-bound of the decoding error $\epsilon$ on the calibration set is given by:
\begin{equation}
    \epsilon_{\mathrm{sup}} = \sum_{i\in S^C}\|\mathbf{\Phi}_{:,i}\|_2^2\|\mathbf{w}_i\|_2^2.
\end{equation}
This lemma indicates that the decoding error upper-bound is related to both the combination coefficients and the basis vectors. To minimize this upper-bound, we should prioritize pruning micro-experts with smaller norms of combination coefficients and basis vectors. The following definition provides a formal description of this intuition.

\paragraph{Definition} The decoding-time energy $\mathcal{E}$ of the \textit{i}-th micro-expert is proportional to the norm of its activation coefficient $\mathbf{\Phi}_{:,i}$ and basis vector $\mathbf{w}_i$, denoted as:
\begin{equation}
    \mathcal{E}_i = \|\mathbf{\Phi}_{:,i}\|_2^2\|\mathbf{w}_i\|_2^2.
\label{eq:energy}
\end{equation}
Based on the definition of energy, we can rank all micro-experts and prioritize retaining the Top-$|S|$ highest-energy ones. Before presenting the specific algorithm, we first provide its tighter error bound.

\paragraph{Theorem} Let $\hat{\mathbf{Y}}$ denote the decoding result using the Top-$|S|$ highest-energy micro-experts, and let $\mathbf{Y}^{*}$ denote the rank-$|S|$ SVD approximation of $\mathbf{Y}$. If $k=N_e-|S|$, the approximation error of $\hat{\mathbf{Y}}$ differs from the optimal SVD by only an $O(k)$-delta, i.e.,
\begin{equation}
    \|\mathbf{Y} - \hat{\mathbf{Y}}\|_F^2 \leq \|\mathbf{Y} - \mathbf{Y}^{*}\|_F^2 + \delta\left(O(k)\right).
\end{equation}
This theorem establishes a relationship with the optimal error bound. The above analysis shows that the energy-based estimation of micro-expert redundancy yields a controllable approximation error. All proofs in this section are given in Section~\ref{subsec:proof} in the Appendix.

Based on this, we propose the \Camera algorithm. Inspired by prior work, we aim to extend the definition of energy by additionally considering the effect of the maximum activation coefficient. The revised computation is given by:
\begin{equation}
    \mathcal{E}_i = \left[\left(1-\alpha\right)\|\mathbf{\Phi}_{:,i}\|_2^2+\alpha\|\mathbf{\Phi}_{:,i}\|_{\infty}^2\right]\cdot\|\mathbf{w}_i\|_2^2.
\label{eq:energy2}
\end{equation}
For a given MoE layer, we estimate the energy of all micro-experts on the calibration dataset using Equation~\ref{eq:energy2} and rank them accordingly. This ranking reflects the importance of each micro-expert: the lower the energy, the more redundant the micro-expert is. Notably, \Camera treats all micro-experts uniformly, without distinguishing which expert they belong to. Moreover, if shared experts exist, they do not need to be treated specially. The procedure for a single MoE layer is summarized in Algorithm~\ref{alg:camera}.

\begin{algorithm}[t!]
\caption{\Camera—Micro-Experts Ranking}
\label{alg:camera}
\textbf{Input}: MoE layer with $N_E$ experts, each with weights $\mathbf{W}_i^\mathrm{up}$, $\mathbf{W}_i^\mathrm{gate}$, $\mathbf{W}_i^\mathrm{down}$; router $\mathbf{R}$; calibration dataset $(\mathbf{X}, \mathbf{Y})$; balance coefficient $\alpha$\\
\textbf{Output}: Rank of all micro-experts
\begin{algorithmic}[1] 
\STATE $\boldsymbol{\phi}^2,\boldsymbol{\phi}^{\infty},\boldsymbol{\omega}^2\gets\texttt{Zeros}(\textrm{shape=}N_E \cdot d_\mathrm{ff})$
\STATE $\mathbf{A}\gets\texttt{Top\&Norm}(\mathbf{RX})$
\FOR {each expert $i = 1$ to $N_E$}
\STATE $p,q \gets i\cdot d_\mathrm{ff},\ (i{+}1)\cdot d_\mathrm{ff}$
\STATE $\boldsymbol{\omega}^2_{p:q} \gets \texttt{ColumnSum}(\mathbf{W}_i^\mathrm{down}\cdot\mathbf{W}_i^\mathrm{down})$
\STATE $\boldsymbol{\Lambda}\in\mathbb{R}^{n\times d_\mathrm{ff}}\gets \mathbf{A}_{:,i}\cdot\sigma\left(\mathbf{W}_i^{\mathrm{gate}}\mathbf{X}\right)\cdot\mathbf{W}_i^{\mathrm{up}}\mathbf{X}$
\STATE $\boldsymbol{\phi}^2_{p:q} \gets \texttt{ColumnSum}(\boldsymbol{\Lambda}\cdot\boldsymbol{\Lambda})$
\STATE $\boldsymbol{\phi}^\infty_{p:q} \gets \texttt{ColumnMax}(\texttt{Abs}(\boldsymbol{\Lambda}))^2$
\ENDFOR
\STATE $\mathcal{E}\gets[(1-\alpha)\boldsymbol{\phi}^2+\alpha\boldsymbol{\phi}^\infty]\cdot\boldsymbol{\omega}^2$
\RETURN $\texttt{Argsort}(\mathcal{E})$
\end{algorithmic}
\end{algorithm}

\begin{algorithm}[t!]
\caption{\CameraP—Micro-Experts Pruning}
\label{alg:camerap}
\textbf{Input}: MoE model $\mathcal{M}$ to be pruned; calibration dataset $\mathcal{C}$; overall pruning ratio $\lambda$\\
\textbf{Output}: Pruned model $\mathcal{M}_\mathrm{prune}$
\begin{algorithmic}[1] 
\STATE $\mathbf{I} \gets \texttt{Embedding}(\mathcal{C})$
\STATE $\mathbf{O} \gets \texttt{ZerosLike}(\mathbf{I})$
\FOR {each layer $L_i \in \{L_1, L_2,\cdots,L_{N_l}\}$}
\STATE $M \gets \texttt{GetMoEModule}(L_i)$
\STATE $h\gets\texttt{RegisterForwardHook}(M)$
\STATE $\mathbf{O} \gets L_i(\mathbf{I})$
\STATE $(\mathbf{X}, \mathbf{Y}) \gets \texttt{GetLayerCalibSamples}(h)$
\STATE $\mathcal{R} \gets$\ \Camera$(M,\mathbf{X},\mathbf{Y},\alpha)$
\STATE $S \gets \texttt{TopSelect}(\mathcal{R}, 1-\lambda)$
\STATE $\mathbf{W}^\mathrm{up} \gets \texttt{Concat}([\mathbf{W}^\mathrm{up}_j\ \text{for}\ j=1\ \text{to}\ N_E])$
\STATE $\mathbf{W}^\mathrm{gate} \gets \texttt{Concat}([\mathbf{W}^\mathrm{gate}_j\ \text{for}\ j=1\ \text{to}\ N_E])$
\STATE $\mathbf{W}^\mathrm{down} \gets \texttt{Concat}([\mathbf{W}^\mathrm{down}_j\ \text{for}\ j=1\ \text{to}\ N_E])$
\STATE $\mathbf{W}^\mathrm{up},\mathbf{W}^\mathrm{gate},\mathbf{W}^\mathrm{down} \gets \mathbf{W}^\mathrm{up}_{S,:},\mathbf{W}^\mathrm{gate}_{S,:},\mathbf{W}^\mathrm{down}_{:,S}$
\STATE $\mathbf{O} \gets L_i(\mathbf{I})$
\STATE $\mathbf{I},\mathbf{O} \gets \mathbf{O},\mathbf{I}$
\ENDFOR
\RETURN $\mathcal{M}_\mathrm{prune}$
\end{algorithmic}
\end{algorithm}

\section{Structured Pruning}

In this section, we propose \CameraP, a multi-matrix joint pruning algorithm for MoE layers based on micro-expert redundancy analysis, along with the associated experiments and analytical results.

\subsection{Pruning Framework \CameraP}

We aim to prune a specified proportion $\lambda$ of micro-experts in each MoE module. Leveraging the \Camera algorithm, 
we identify the less important micro-experts in each layer.
Specifically, for each selected micro-expert, all three associated weight vectors are simultaneously zeroed out. The detailed procedure is outlined in Algorithm~\ref{alg:camerap}, which applies pruning layer by layer. First, lines $5\sim7$ collect calibration samples from the MoE module, including the input and output hidden states. Then, the \Camera greedy algorithm is used to rank all micro-experts and determine the set $S$ of those to retain. Within each expert, only the micro-experts with indices in $S$ are preserved. Finally, the post-pruned output $\mathbf{O}$ is recomputed, 
proceeding to the next layer.

\begin{table*}[t!]
\centering
\small
\begin{tabular}{c@{\hspace{0.20cm}}|@{\hspace{0.40cm}}c@{\hspace{0.40cm}}|c@{\hspace{0.6cm}}c|c@{\hspace{0.30cm}}c@{\hspace{0.30cm}}c@{\hspace{0.30cm}}c@{\hspace{0.30cm}}c@{\hspace{0.30cm}}c@{\hspace{0.30cm}}c@{\hspace{0.30cm}}c@{\hspace{0.30cm}}c@{\hspace{0.30cm}}c}
\toprule
$\lambda$ & Method & Wiki2 & C4 & BoolQ & OBQA & RTE & Wino. & Hella. & PIQA & MathQA & ARC-e & ARC-c & Avg. \\
\midrule
\multicolumn{14}{c}{\textbf{Deepseek-MoE-16B-base}\quad(2 shared experts + 64 common experts, 2+top-4 experts are activated)}\\
\cmidrule(lr){1-14}
0\% & Original & 6.51 & 9.05 & 72.45 & 44.00 & 63.54 & 70.24 & 77.37 & 80.68 & 31.52 & 72.89 & 47.86 & 62.28 \\
\cmidrule(lr){1-14}
\multirow{3}{*}[-0ex]{20\%} & NAEE & 6.77 & 10.07 & 67.83 & 42.40 & 62.09 & 69.53 & 74.63 & 78.34 & 30.99 & \textbf{72.56} & \textbf{46.24} & 60.51 \\
& $D^2$-MoE & 7.29 & 12.62 & \textbf{69.32} & 41.40 & 61.01 & 69.22 & 69.87 & 76.44 & 29.45 & 71.29 & 42.75 & 58.97 \\
& \CameraP & \textbf{6.57} & \textbf{9.84} & 68.01 & \textbf{44.00} & \textbf{64.62} & \textbf{70.17} & \textbf{75.02} & \textbf{78.62} & \textbf{31.46} & 71.80 & 45.56 & \textbf{61.03} \\
\cmidrule(lr){1-14}
\multirow{3}{*}[-0ex]{40\%} & NAEE & 8.01 & 12.80 & 62.26 & 39.60 & 57.40 & 63.69 & 66.16 & 75.41 & 27.37 & 64.06 & 38.48 & 54.94 \\
& $D^2$-MoE & 8.38 & 17.22 & 66.05 & 36.60 & 57.03 & 66.77 & 58.74 & 71.44 & 27.67 & 66.03 & 38.57 & 54.32 \\
& \CameraP & \textbf{6.93} & \textbf{11.68} & \textbf{70.64} & \textbf{43.20} & \textbf{58.48} & \textbf{68.51} & \textbf{69.04} & \textbf{75.41} & \textbf{29.01} & \textbf{70.71} & \textbf{42.24} & \textbf{58.58} \\
\cmidrule(lr){1-14}
\multirow{3}{*}[-0ex]{60\%} & NAEE & 15.47 & 29.44 & 51.65 & 30.60 & \textbf{58.48} & 53.67 & 47.50 & 65.88 & 22.31 & 48.90 & 28.50 & 45.28 \\
& $D^2$-MoE & 12.13 & 34.54 & 61.78 & 31.60 & 53.43 & 61.09 & 43.29 & 63.87 & 23.65 & 50.59 & 31.14 & 46.72 \\
& \CameraP & \textbf{8.68} & \textbf{18.10} & \textbf{62.60} & \textbf{40.20} & 56.32 & \textbf{64.33} & \textbf{56.53} & \textbf{67.90} & \textbf{26.16} & \textbf{54.88} & \textbf{35.67} & \textbf{51.62} \\
\midrule
\multicolumn{14}{c}{\textbf{Qwen2-57B-A14B}\quad(8 shared experts + 64 common experts, 8+top-8 experts are activated)}\\
\cmidrule(lr){1-14}
0\% & Original & 5.92 & 8.22 & 86.39 & 44.20 & 75.45 & 73.48 & 82.56 & 81.61 & 38.22 & 69.53 & 49.23 & 66.74 \\
\cmidrule(lr){1-14}
\multirow{3}{*}[-0ex]{20\%} & NAEE & 6.32 & 8.87 & 86.08 & 43.80 & 74.73 & 73.79 & 81.19 & \textbf{81.55} & 35.14 & 69.36 & 49.32 & 66.11 \\
& $D^2$-MoE & 6.15 & 9.66 & \textbf{87.21} & 44.00 & 74.01 & \textbf{75.37} & 79.94 & 80.63 & 39.00 & 69.53 & 47.70 & 66.38 \\
& \CameraP & \textbf{6.03} & \textbf{8.73} & 85.57 & \textbf{45.20} & \textbf{74.73} & 74.03 & \textbf{82.02} & 81.18 & \textbf{41.71} & \textbf{70.37} & \textbf{50.68} & \textbf{67.28} \\
\cmidrule(lr){1-14}
\multirow{3}{*}[-0ex]{40\%} & NAEE & 7.72 & 10.56 & 82.20 & 42.00 & 69.68 & 70.88 & 76.68 & 79.81 & 34.04 & 70.20 & 49.83 & 63.92 \\
& $D^2$-MoE & 6.37 & 12.74 & 85.50 & 42.40 & 74.62 & 72.93 & 72.25 & 76.99 & 35.48 & 70.58 & 48.81 & 64.40 \\
& \CameraP & \textbf{6.31} & \textbf{9.75} & \textbf{86.42} & \textbf{45.40} & \textbf{74.73} & \textbf{73.88} & \textbf{80.34} & \textbf{80.03} & \textbf{36.85} & \textbf{73.10} & \textbf{50.51} & \textbf{66.81} \\
\cmidrule(lr){1-14}
\multirow{3}{*}[-0ex]{60\%} & NAEE & 16.68 & 21.69 & 66.64 & 34.20 & 56.68 & 62.04 & 57.88 & 70.24 & 25.06 & 55.89 & 33.96 & 51.40 \\
& $D^2$-MoE & 11.56 & 25.92 & 75.69 & 35.40 & 72.56 & 70.80 & 57.35 & 70.35 & 27.10 & 59.22 & 38.40 & 56.32 \\
& \CameraP & \textbf{7.24} & \textbf{12.45} & \textbf{83.79} & \textbf{42.40} & \textbf{78.70} & \textbf{72.38} & \textbf{73.23} & \textbf{76.39} & \textbf{35.78} & \textbf{73.11} & \textbf{50.77} & \textbf{65.17} \\
\midrule
\multicolumn{14}{c}{\textbf{Qwen3-30B-A3B}\quad(128 common experts, top-8 experts are activated)}\\
\cmidrule(lr){1-14}
0\% & Original & 8.70 & 12.15 & 88.69 & 44.40 & 81.23 & 70.64 & 77.63 & 80.52 & 59.20 & 79.08 & 56.31 & 70.86 \\
\cmidrule(lr){1-14}
\multirow{3}{*}[-0ex]{20\%} & NAEE & 8.72 & 12.44 & \textbf{88.74} & 44.40 & \textbf{83.39} & 69.85 & 77.32 & 80.14 & 49.27 & 77.31 & \textbf{56.31} & 69.64 \\
& $D^2$-MoE & 9.68 & 18.52 & 85.90 & 42.80 & 80.87 & 68.19 & 74.26 & 78.40 & 48.81 & 71.46 & 46.50 & 66.35 \\
& \CameraP & \textbf{8.48} & \textbf{12.25} & 88.50 & \textbf{44.40} & 82.67 & \textbf{70.56} & \textbf{77.51} & \textbf{80.30} & \textbf{52.80} & \textbf{78.33} & 54.43 & \textbf{69.94} \\
\cmidrule(lr){1-14}
\multirow{3}{*}[-0ex]{40\%} & NAEE & 9.29 & \textbf{13.87} & \textbf{87.21} & 43.00 & 74.01 & \textbf{68.74} & 73.50 & 77.74 & 42.91 & 72.72 & 52.13 & 65.77 \\
& $D^2$-MoE & 20.40 & 35.48 & 85.50 & 42.00 & 76.53 & 64.33 & 69.87 & 71.44 & 40.90 & 69.78 & 44.62 & 62.77 \\
& \CameraP & \textbf{9.29} & 15.58 & 86.97 & \textbf{43.00} & \textbf{79.42} & 67.64 & \textbf{73.79} & \textbf{78.51} & \textbf{44.96} & \textbf{77.31} & \textbf{55.55} & \textbf{67.46} \\
\cmidrule(lr){1-14}
\multirow{3}{*}[-0ex]{60\%} & NAEE & \textbf{12.08} & \textbf{19.37} & 72.23 & 36.60 & 68.95 & 63.93 & 62.06 & 71.87 & 28.68 & \textbf{66.92} & 42.92 & 57.13 \\
& $D^2$-MoE & 32.13 & 68.36 & 70.64 & 35.40 & 64.62 & 59.12 & 58.21 & 63.23 & 26.00 & 57.37 & 33.57 & 52.02 \\
& \CameraP & 12.48 & 24.48 & \textbf{82.08} & \textbf{41.00} & \textbf{68.95} & \textbf{64.64} & \textbf{64.90} & \textbf{73.01} & \textbf{30.89} & 62.46 & \textbf{43.35} & \textbf{59.03} \\
\bottomrule
\end{tabular}
\caption{Main pruning results of evaluation experiment on three mainstream MoE models. The best scores are in bold. We also test the performance of the original model (16-bit) as a reference.}
\label{tab:camerap}
\end{table*}

\begin{table*}[t!]
\centering
\small
\begin{tabular}{c@{\hspace{0.20cm}}|@{\hspace{0.20cm}}c@{\hspace{0.20cm}}|c@{\hspace{0.6cm}}c|c@{\hspace{0.30cm}}c@{\hspace{0.30cm}}c@{\hspace{0.30cm}}c@{\hspace{0.30cm}}c@{\hspace{0.30cm}}c@{\hspace{0.30cm}}c@{\hspace{0.30cm}}c@{\hspace{0.30cm}}c@{\hspace{0.30cm}}c}
\toprule
Wbits & Method & Wiki2 & C4 & BoolQ & OBQA & RTE & Wino. & Hella. & PIQA & MathQA & ARC-e & ARC-c & Avg. \\
\midrule
\multicolumn{14}{c}{\textbf{Deepseek-MoE-16B-base}\quad(2 shared experts + 64 common experts, 2+top-4 experts are activated)}\\
\cmidrule(lr){1-14}
16-bit & Original & 6.51 & 9.05 & 72.45 & 44.00 & 63.54 & 70.24 & 77.37 & 80.68 & 31.52 & 72.89 & 47.86 & 62.28 \\
\cmidrule(lr){1-14}
\multirow{4}{*}[-0ex]{2.25-bit} & GPTQ & 11.36 & 14.34 & 61.44 & 37.60 & 55.23 & 64.01 & 64.35 & 75.57 & 25.23 & 62.25 & 35.41 & 53.45 \\
& MC & 11.10 & 16.54 & 60.82 & 38.80 & \textbf{56.32} & 64.25 & 67.77 & 76.33 & 24.69 & 62.46 & 38.65 & 54.45 \\
& \CameraQd & 11.28 & 14.70 & 61.68 & 38.40 & 53.60 & 64.33 & 61.45 & 76.88 & 26.33 & 59.04 & 32.50 & 52.69 \\
& \CameraQ & \textbf{9.51} & \textbf{12.26} & \textbf{66.94} & \textbf{39.00} & 55.23 & \textbf{66.46} & \textbf{69.21} & \textbf{77.04} & \textbf{28.07} & \textbf{66.54} & \textbf{40.52} & \textbf{56.56} \\
\midrule
\multicolumn{14}{c}{\textbf{Qwen3-30B-A3B}\quad(128 common experts, top-8 experts are activated)}\\
\cmidrule(lr){1-14}
16-bit & Original & 8.70 & 12.15 & 88.69 & 44.40 & 81.23 & 70.64 & 77.63 & 80.52 & 59.20 & 79.08 & 56.31 & 70.85 \\
\cmidrule(lr){1-14}
\multirow{4}{*}[-0ex]{2.25-bit} & GPTQ & 13.72 & 16.06 & 72.51 & 33.80 & 68.59 & 59.83 & 66.04 & 70.62 & 24.36 & 46.89 & 32.17 & 52.76 \\
& MC & 13.59 & 15.20 & \textbf{72.57} & 34.20 & \textbf{68.95} & 61.08 & 68.30 & 68.28 & 25.06 & 45.49 & 31.14 & 52.79 \\
& \CameraQd & 14.06 & 15.79 & 71.58 & 34.20 & 67.15 & 62.04 & 67.14 & 71.44 & 24.52 & 45.90 & 28.82 & 52.53 \\
& \CameraQ & \textbf{12.13} & \textbf{14.97} & 72.32 & \textbf{36.80} & 64.62 & \textbf{62.75} & \textbf{71.19} & \textbf{74.16} & \textbf{26.00} & \textbf{50.38} & \textbf{34.56} & \textbf{54.75} \\
\bottomrule
\end{tabular}
\caption{Main mixed-precision quantization results of evaluation experiment. The best scores are in bold.}
\label{tab:cameraq}
\end{table*}

\begin{algorithm}[t]
\caption{\CameraQ—Mixed-precision Quantization}
\label{alg:cameraq}
\textbf{Input}: MoE model $\mathcal{M}$ to be quantized; calibration dataset $\mathcal{C}$; ratio list $\{r_1,r_2,r_3\}$; bit-width list $\{b_1,b_2,b_3\}$\\
\textbf{Output}: Quantized model $\mathcal{M}_\mathrm{quant}$
\begin{algorithmic}[1] 
\STATE Get initial $\mathbf{I},\mathbf{O}$ such as in \CameraP
\FOR {each layer $L_i \in \{L_1, L_2,\cdots,L_{N_l}\}$}
\STATE Get calib-samples $(\mathbf{X},\mathbf{Y})$ such as in \CameraP
\STATE $\mathcal{R} \gets$\ \Camera$(M,\mathbf{X},\mathbf{Y},\alpha)$
\STATE $S_1,S_2,S_3 \gets \texttt{ListSplitByRatio}(\mathcal{R},r_1,r_2,r_3)$
\FOR {each expert $j = 1$ to $N_E$}
\STATE $p,q \gets j\cdot d_\mathrm{ff},\ (j{+}1)\cdot d_\mathrm{ff}$
\STATE $s_1,s_2,s_3\gets \texttt{GetSubIndex}(S_1,S_2,S_3,[p,q])$
\FOR {each bit-width $b_k\in\{b_1,b_2,b_3\}$}
\STATE $\mathbf{W}_{s_k,:}^\mathrm{up} \gets \texttt{Quantize}(\mathbf{W}_{s_k,:}^\mathrm{up},b_k)$
\STATE $\mathbf{W}_{s_k,:}^\mathrm{gate} \gets \texttt{Quantize}(\mathbf{W}_{s_k,:}^\mathrm{gate},b_k)$
\STATE $\mathbf{W}_{,:s_k}^\mathrm{down} \gets \texttt{Quantize}(\mathbf{W}_{,:s_k}^\mathrm{down},b_k)$
\ENDFOR
\ENDFOR
\STATE Recompute $\mathbf{I},\mathbf{O}$ such as in \CameraP
\ENDFOR
\RETURN $\mathcal{M}_\mathrm{quant}$
\end{algorithmic}
\end{algorithm}

\subsection{Experimental Setup}

\paragraph{Models and Data} We evaluate our method on three modern MoE models with more, smaller experts: Deepseek-MoE-16B, Qwen2-57B-A14B, and Qwen3-30B-A3B. Each is tested under 20\%-60\% overall pruning ratios. Earlier, fewer-expert models are discussed in Appendix~\ref{subsec:supexp}. All methods use Wikitext2 for calibration; ours uses 128 sequences of 2048 tokens. Other settings are in Appendix~\ref{subsec:details}.

\paragraph{Baselines} We compare \CameraP with two strong baselines: NAEE~\citep{naee2024}, which prunes entire experts directly, and $D^2$-MoE~\citep{delta2025}, a recent strong merge-then-compress approach. To address the scalability issue with $N_E \geq 8$ in NAEE, we follow MoE-$I^2$~\cite{moei2024} and adopt efficient genetic search. \CameraP uses $\alpha=0.95,0.95,1.00$ for the three models.

\paragraph{Evaluation} To evaluate baseline performance, we calculate perplexity on randomly sampled sequences from Wikitext2~\citep{wiki22016} and C4~\citep{C42020}—lower is better. We also report accuracy on nine zero-shot downstream tasks, including Winogrande~\citep{winogrande2021}, HellaSwag~\citep{hellaswag2019}, PIQA~\citep{piqa2020}, BoolQ~\citep{boolq2019}, ARC-e/ARC-c~\citep{arc2018}, OBQA~\citep{obqa2018}, MathQA~\citep{mathqa2019} and RTE~\citep{glue2018}. We prioritize outputting \texttt{acc\_norm} from LM-Evaluation-Harness\footnote{\url{https://github.com/EleutherAI/lm-evaluation-harness}}.

\subsection{Main Results}

We evaluate both shared and non-shared expert models, as shown in Table~\ref{tab:camerap}. Across all models and pruning ratios, our method consistently outperforms strong baselines in both perplexity and accuracy, especially under higher pruning rates. $D^2$-MoE often ranks second, except in certain Qwen3 configurations. However, its SVD-based approach suffers from numerical instability in several Qwen2/3 layers, while our method remains stable. We believe the superior performance of \CameraP stems from preserving the functional structure of micro-experts across matrices, maintaining important ones with full precision. Moreover, \CameraP is a training- and gradient-free method that completes in just 0.1 GPU hours, making it over 100x faster than competing methods. As for NAEE, this brute-force combinatorial approximation seems beneficial only at low pruning ratios.

\section{Mixed-precision Quantization}

In this section, we propose \CameraQ, a cross-expert mixed-precision bit allocation strategy for MoE models, and present the corresponding experimental results. Notably, \CameraQ is not introduced as a standalone quantization algorithm, but rather as a complementary component that can be \textbf{integrated} with any existing weight quantization method to enable mixed-precision quantization.

\subsection{Mixed-precision Strategy \CameraQ}

We illustrate \CameraQ in a setting with three predefined precision levels, where each matrix is partitioned into three segments. As shown in line 4 of Algorithm~\ref{alg:cameraq}, \Camera provides a global importance ranking of all micro-experts, forming the basis for mixed-precision assignment. Line 5 then divides this ranking into index sets $S_i$, each corresponding to a different precision level. Notably, these sets span all experts across the entire MoE layer. For each expert, we extract the indices $s_i$ of its micro-experts within each $S_i$, and quantize the corresponding sub-matrices accordingly. Lines from 10 to 12 handle rows and columns differently to ensure consistent precision within each micro-expert. Additionally, micro-experts in each expert are reordered beforehand so that those assigned the same precision level are colocated.

\subsection{Experimental Setup}

We perform 2-bit mixed-precision quantization on MoE layers on Deepseek-MoE-16B and Qwen3-30B-A3B, focusing on comparisons with single-precision GPTQ~\citep{gptq2022} and the recent MC method~\citep{mc2025} that applies mixed precision at the expert level. For fairness, \CameraQ also adopts GPTQ for its quantization step (Lines 10$\sim$12 in Algorithm~\ref{alg:cameraq}). We further introduce a variant baseline, \CameraQd, which applies mixed precision by slicing all matrices along the input dimension (i.e., column-wise), following the most common practice. Specifically, our method is $\mathbf{W}_{s_k,:}^\mathrm{up},\mathbf{W}_{s_k,:}^\mathrm{gate},\mathbf{W}_{:,s_k}^\mathrm{down}$, while \CameraQd uses $\mathbf{W}_{:,s_k}^\mathrm{up},\mathbf{W}_{:,s_k}^\mathrm{gate},\mathbf{W}_{:,s_k}^\mathrm{down}$, breaking precision consistency within each micro-expert. We analyze this difference in detail in Section~\ref{subsec:integ}. Precision settings are in Appendix~\ref{subsec:mixsetting}. All baselines are calibrated on the C4 dataset, and our method uses 128 sequences of length 2048. Evaluation follows the same setting as \CameraP.

\begin{figure*}[t]
\centering
\includegraphics[width=1.00\textwidth]{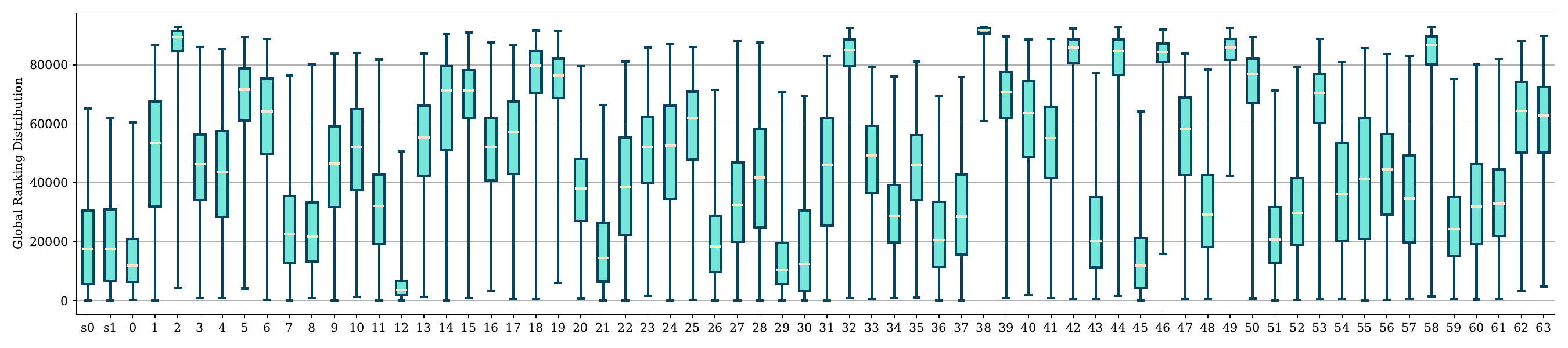}
\caption{Distribution of micro-experts within each expert based on global ranking from \Camera ($\lambda=20\%$), layer 12 of Deepseek-MoE-16B. We list all 66 experts, where `S0/S1' denotes the shared experts, and the rest are non-shared experts.}
\label{fig:dist}
\end{figure*}

\subsection{Main Results}

As shown in Table~\ref{tab:cameraq}, \CameraQ demonstrates clear superiority over other baselines on both perplexity and accuracy. GPTQ serves as a standard baseline for 2-bit single-precision quantization, while MC achieves slightly better results by adjusting expert-level bit allocations based on activation frequency and pre-quantization loss. However, we stress that such coarse-grained expert-level allocation fails to capture the finer differences among micro-experts. \CameraQ addresses this limitation, and its strong performance supports the effectiveness of a more fine-grained approach. Moreover, \CameraQ adopts a matrix partitioning distinct from that used in previous methods, which also contributes to its performance gains, as evidenced by the performance drop observed in \CameraQd.

\begin{figure}[t]
\centering
\includegraphics[width=0.99\columnwidth]{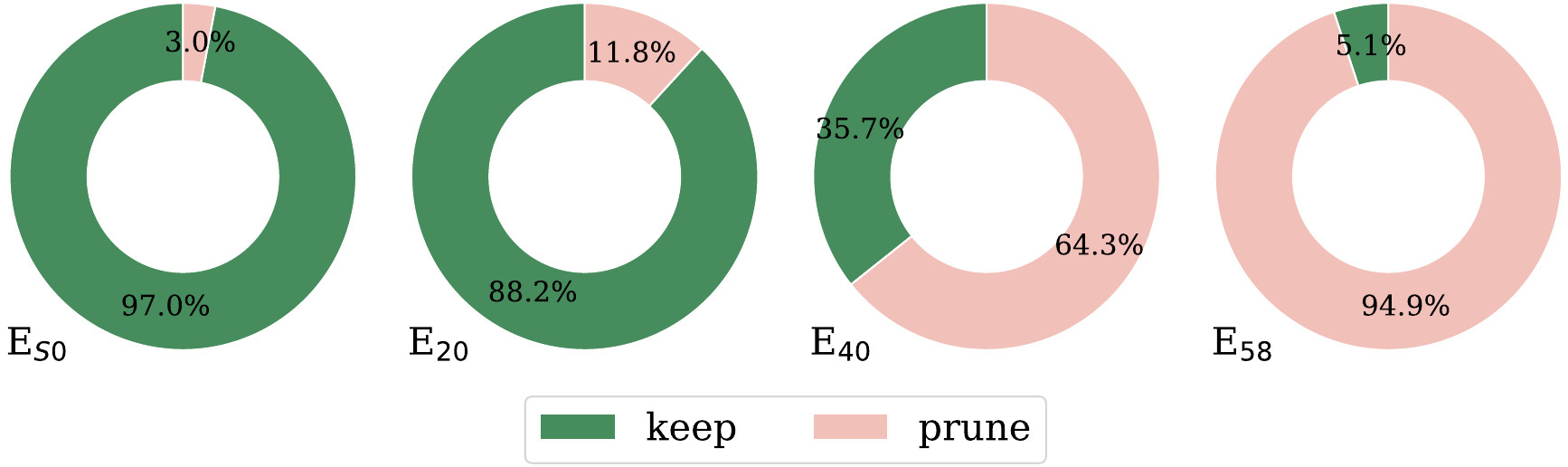}
\caption{Pruning ratios across selected experts, taken from layer 12 of Deepseek-MoE-16B, with $\lambda = 40\%$.}
\label{fig:cut}
\end{figure}

\begin{figure}[t]
\centering
\includegraphics[width=0.99\columnwidth]{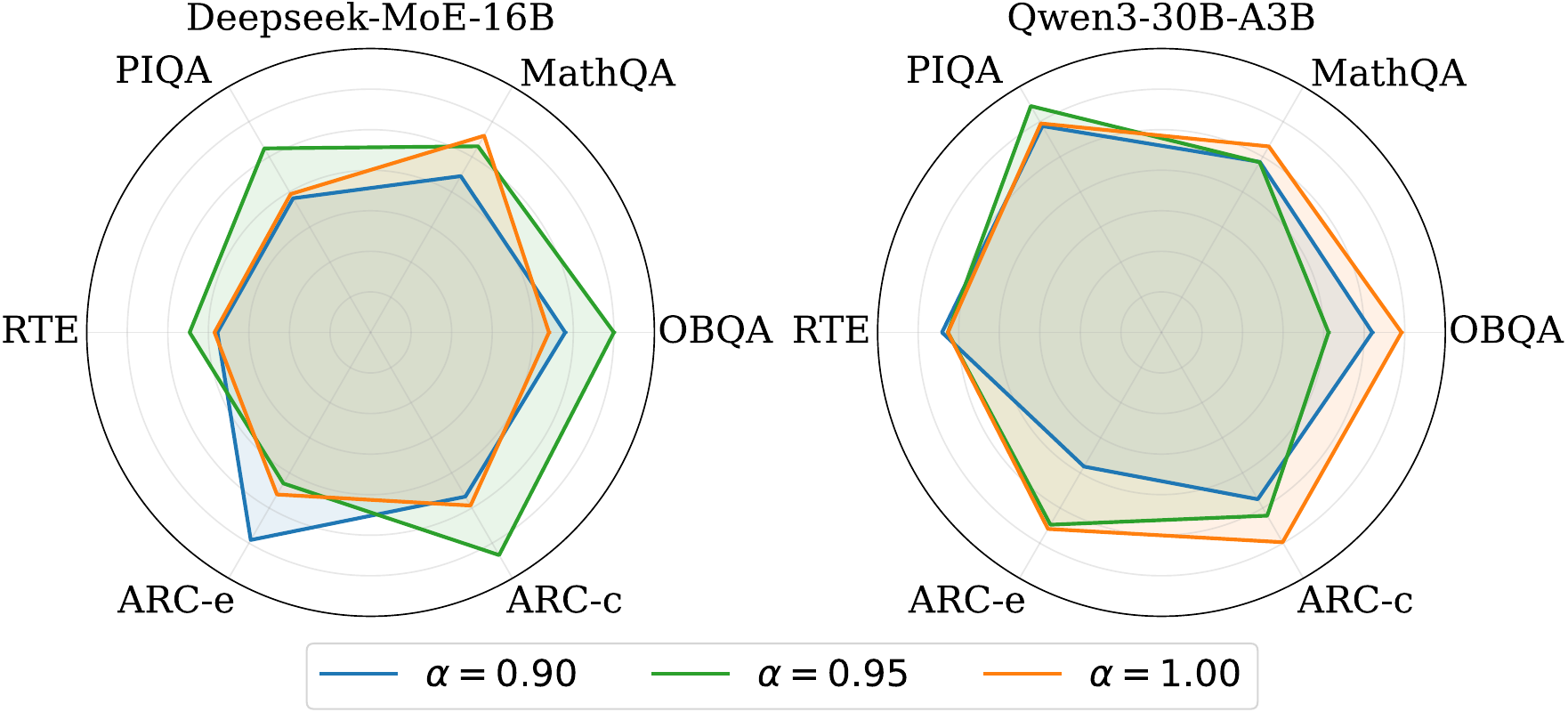}
\caption{Task performance with varying $\alpha$ when $\lambda=20\%$. The scores are scaled to highlight the differences.}
\label{fig:coef}
\end{figure}

\begin{table}[t]
\centering
\small
\begin{tabular}{c@{\hspace{0.20cm}}|@{\hspace{0.20cm}}c@{\hspace{0.20cm}}c@{\hspace{0.20cm}}c@{\hspace{0.20cm}}c@{\hspace{0.20cm}}c@{\hspace{0.20cm}}c@{\hspace{0.20cm}}}
\toprule
$|\mathcal{C}|$ & 8 & 16 & 32 & 64 & 128 & 256 \\
\midrule
Wikitext2 & 6.66 & 6.60 & 6.59 & 6.57 & 6.57 & 6.56 \\
C4 & 7.28 & 7.13 & 7.07 & 6.94 & 6.95 & 6.92 \\
\bottomrule
\end{tabular}
\caption{Perplexity on Wikitext2 with different sizes and sources of calibration data. $|\mathcal{C}|$ denotes the size of $\mathcal{C}$.}
\label{tab:calib}
\end{table}

\begin{table}[t]
\centering
\small
\begin{tabular}{c@{\hspace{0.20cm}}|@{\hspace{0.20cm}}c@{\hspace{0.20cm}}c@{\hspace{0.20cm}}c@{\hspace{0.20cm}}}
\toprule
Method & NAEE & $D^2$-MoE & \CameraP \\
\midrule
$\lambda=20\%$ & 1.00x & 1.03x & 1.06x \\
$\lambda=40\%$ & 1.00x & 1.04x & 1.42x \\
$\lambda=60\%$ & 1.00x & 1.08x & 1.48x \\
\bottomrule
\end{tabular}
\caption{Decoding speed comparison of pruned MoE layer in Deepseek-MoE-16B. The batch size is 64 tokens.}
\label{tab:speed}
\end{table}

\section{Discussion}

\subsection{Micro-Expert Distribution}
\label{subsec:medis}
Figure~\ref{fig:dist} illustrates the significant variation in the distribution of micro-experts across different experts within the same layer. A lower rank indicates higher importance. Notably, the two shared experts, along with experts 0, 12, 29, and 45, stand out as particularly important, as most of their micro-experts rank highly at the global level. This provides strong evidence supporting the core assumption of \Camera and underscores the advantage of fine-grained expert pruning over coarse-grained methods, as shown in Figure~\ref{fig:cut}.

\subsection{Approximation Errors}

Our better performance is interpretable. By retaining the most key micro-experts, the outputs of the pruned MoE layers under \CameraP remain closer to those of the original model—both in terms of L2 distance and cosine similarity. Please refer to Appendix~\ref{subsec:apperr} for detailed results.

\subsection{Ablation on Balance Coefficient}

The $\alpha$ in Equation~\ref{eq:energy2} slightly affects the ranking of micro-experts and downstream performance, but has little impact on perplexity or average accuracy. Figure~\ref{fig:coef} shows how different $\alpha$ influences task scores. We infer that task- or domain-specific micro-experts lead to this behavior. In experiments, we choose the $\alpha$ that produced the highest scores.

\begin{figure*}[t]
\centering
\includegraphics[width=0.95\textwidth]{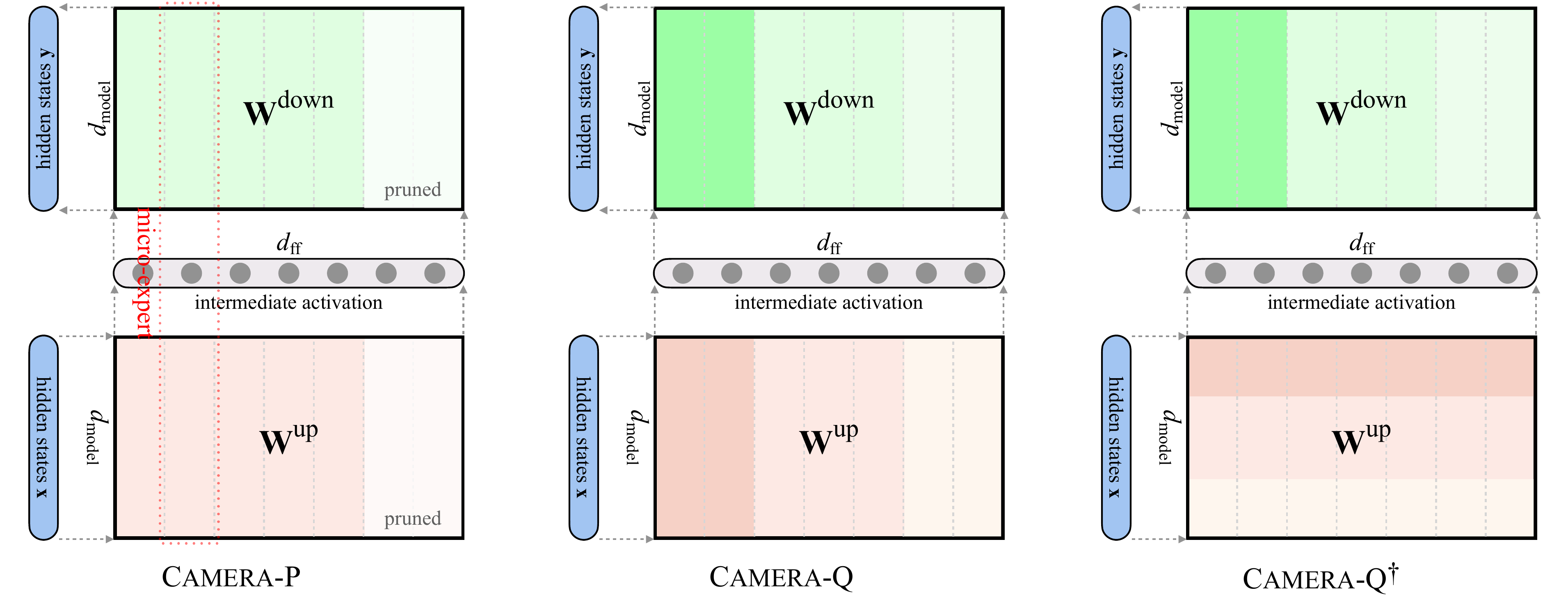}
\caption{Matrix calculation flow of \CameraP, \CameraQ and \CameraQd. For simplicity, we omit the matrix $\mathbf{W^\mathrm{gate}}$. The red dashed box on the left indicates the weight of a micro-expert. In \CameraQ and \CameraQd, we use light and dark colors to indicate the lower and higher bit-width of the weights.}
\label{fig:comparison}
\end{figure*}

\begin{table*}[t]
\centering
\small
\begin{tabular}{@{\hspace{0.20cm}}c@{\hspace{0.20cm}}|c@{\hspace{0.6cm}}c|c@{\hspace{0.30cm}}c@{\hspace{0.30cm}}c@{\hspace{0.30cm}}c@{\hspace{0.30cm}}c@{\hspace{0.30cm}}c@{\hspace{0.30cm}}c@{\hspace{0.30cm}}c@{\hspace{0.30cm}}c@{\hspace{0.30cm}}c}
\toprule
Method & Wiki2 & C4 & BoolQ & OBQA & RTE & Wino. & Hella. & PIQA & MathQA & ARC-e & ARC-c & Avg. \\
\midrule
\multicolumn{13}{c}{\textbf{openPangu Embedded-7B-V1.1}\quad(dense model, only 1 expert)}\\
\cmidrule(lr){1-13}
Original (16-bit) & 34.94 & 43.73 & 77.74 & 29.80 & 67.87 & 55.25 & 59.98 & 68.72 & 26.20 & 54.97 & 35.49 & 52.89 \\
\cmidrule(lr){1-13}
\CameraP (20\%) & 44.35 & 48.67 & 75.20 & 31.00 & 65.34 & 54.14 & 54.79 & 66.43 & 23.75 & 52.90 & 35.41 & 51.00 \\
+Wanda (20\%+20\%) & 42.96 & 48.11 & 74.56 & 30.80 & 66.06 & 55.09 & 54.34 & 67.19 & 24.09 & 53.07 & 34.81 & 51.11 \\
\bottomrule
\end{tabular}
\caption{Results of \CameraP and integrating Wanda. ``20\%+20\%'' denotes a pruning ratio of ``20\%+80\%$\times$20\%=36\%''.}
\label{tab:pangu}
\end{table*}

\subsection{Ablation on Calibration Dataset}

\Camera achieves strong performance using only a small amount of general-domain data. Table~\ref{tab:calib} shows that \CameraP is insensitive to both the source and size of calibration data when pruning 20\% of Deepseek-MoE-16B.

\subsection{Decoding Speed on Pruned MoE Layer}

NAEE does not reduce weights during decoding, and while $D^2$-MoE performs rank reduction, it requires additional steps. In contrast, \CameraP directly reduces weights, enabling more efficient decoding, as shown in Table~\ref{tab:speed}.

\subsection{Integrity of Micro-Experts}
\label{subsec:integ}

A central insight of our method is that jointly compressing multiple matrices preserves the functional integrity of micro-experts. Figure~\ref{fig:comparison} compares three different methods. All share the same FFN structure: the input $\mathbf{x}$ is transformed by $\mathbf{W}^\mathrm{up}$ and $\mathbf{W}^\mathrm{gate}$ into intermediate activations, followed by $\mathbf{W}^\mathrm{down}$ to produce $\mathbf{y}$. As highlighted in the red dashed box, each row of $\mathbf{W}^\mathrm{up}$ maps uniquely to a column of $\mathbf{W}^\mathrm{down}$, forming a one-to-one micro-expert. For visualization, $\mathbf{W}^\mathrm{up}$ is shown transposed and micro-experts are energy-ranked from left to right.

\CameraP prunes low-energy micro-experts while keeping the remains intact. \CameraQ assigns higher precision to higher-energy micro-experts, enforcing uniform precision within each. In contrast, \CameraQd slices $\mathbf{W}^\mathrm{up}$ and $\mathbf{W}^\mathrm{gate}$ along an orthogonal dimension, quantizing each weight with multiple precisions. Like many existing mixed-precision approaches, \CameraQd operates on individual matrices and allocates bits along the input dimension (e.g., using $\mathbf{H}=\mathbf{XX}^\mathrm{T}$), overlooking cross-matrix expert structure. This often gives high-energy micro-experts inconsistent or insufficient precision. Our experiments show that enforcing precision consistency within each micro-expert, as in \CameraQ, yields substantially better performance.

\subsection{Application on Dense Model}
\label{subsec:dense}

\Camera investigates redundancy that spans across matrices within MoE layers. After applying cross-matrix compression, the resulting MoE layers can still be further compressed using single-matrix methods such as Wanda~\cite{wanda2024}. Moreover, our approach can also be directly applied to structured pruning or mixed-precision quantization in dense-architecture models. As an illustrative example, we apply \CameraP to prune 20\% of the FFN layers in the \textit{openPangu-7B} model~\cite{pangu2025} running on \textit{Ascend 910B}. On top of this, we further apply Wanda to perform an additional 20\% unstructured pruning. The resulting performance is summarized in Table~\ref{tab:pangu}. These results demonstrate that expert-level pruning and intra-matrix pruning can be seamlessly combined to achieve higher overall pruning ratios while maintaining lossless performance.

\section{Conclusion}


We propose \CameraP, a novel, effective, and efficient MoE pruning method grounded in a cross-matrix perspective of micro-experts and guided by the \Camera ranking algorithm. We further highlight the importance of micro-expert-oriented mixed-precision idea in \CameraQ.

\section*{Acknowledgement}

We gratefully acknowledge the support of the National Natural Science Foundation of China (NSFC) via grants 62236004, 62206078 and 62476073. Additional gratitude is extended to our collaborators and colleagues for their insightful discussions during the development of this work.

\section*{Appendix}

According to the AAAI Proceedings policy, appendices and supplementary materials are not included in the published proceedings. We therefore provide the complete version of the paper, including all appendices, on arXiv for reference of readers. Please refer to \url{https://arxiv.org/abs/2508.02322} for the complete appendix. 

\bibliography{aaai2026}

\clearpage

\onecolumn
\appendix

\section{Appendix}

\subsection{Derivation of Micro-Expert}
\label{subsec:derive}

In Section~\ref{subsec:me}, we provide a loose definition of micro-experts by splitting the internal weight matrices of each expert. In this part, we derive a more rigorous definition of a mixture of micro-experts directly from the MoE formulation. All notations follow the same meanings as in the main text. For an MoE layer with $N_E$ experts, we formalize its mathematical definition based on Equation~\ref{eq:E}:
\begin{align}
\mathbf{y} &= \sum_i^{N_E} A_i\left(\mathbf{x}\right)\cdot E_i\left(\mathbf{x}\right)\notag\\
&=\sum_i^{N_E}A_i\cdot \mathbf{W}_i^{\mathrm{down}}\left[ \sigma\left(\mathbf{W}_i^{\mathrm{gate}}\mathbf{x}\right)\cdot\mathbf{W}_i^{\mathrm{up}}\mathbf{x} \right].\notag
\end{align}
The results of the matrix-vector multiplications $\mathbf{W}_i^{\mathrm{up}}\mathbf{x}$ and $\mathbf{W}_i^{\mathrm{gate}}\mathbf{x}$ are vectors of dimension $d_\mathrm{ff}$, as shown below:
$$\mathbf{W}_i^{\mathrm{gate}}\mathbf{x}=\left(\mathbf{W}_i^{\mathrm{gate}}[0,:]\mathbf{x},\mathbf{W}_i^{\mathrm{gate}}[1,:]\mathbf{x},\cdots,\mathbf{W}_i^{\mathrm{gate}}[d_\mathrm{ff},:]\mathbf{x}\right),$$
$$\mathbf{W}_i^{\mathrm{up}}\mathbf{x}=\left(\mathbf{W}_i^{\mathrm{up}}[0,:]\mathbf{x},\mathbf{W}_i^{\mathrm{up}}[1,:]\mathbf{x},\cdots,\mathbf{W}_i^{\mathrm{up}}[d_\mathrm{ff},:]\mathbf{x}\right).$$
The element-wise multiplication of two given vectors is:
$$\sigma\left(\mathbf{W}_i^{\mathrm{gate}}\mathbf{x}\right)\cdot\mathbf{W}_i^{\mathrm{up}}\mathbf{x}=\left[\sigma\left(\mathbf{W}_i^{\mathrm{gate}}[0,:]\mathbf{x}\right)\cdot\mathbf{W}_i^{\mathrm{up}}[0,:]\mathbf{x},\cdots,\sigma\left(\mathbf{W}_i^{\mathrm{gate}}[d_\mathrm{ff},:]\mathbf{x}\right)\cdot\mathbf{W}_i^{\mathrm{up}}[d_\mathrm{ff},:]\mathbf{x}\right].$$
The product of matrix $\mathbf{W}_i^\mathrm{down}$ and the vector above can be viewed as a linear combination of the column vectors of $\mathbf{W}_i^\mathrm{down}$, with the entries of the vector serving as the combination coefficients. Thus, we have:
\begin{align}
\mathbf{y} &=\sum_i^{N_E}A_i\cdot \sum_j^{d_\mathrm{ff}}\sigma\left(\mathbf{W}_i^{\mathrm{gate}}[j,:]\mathbf{x}\right)\cdot\mathbf{W}_i^{\mathrm{up}}[j,:]\mathbf{x}\cdot\mathbf{W}_i^{\mathrm{down}}[:,j].\notag
\end{align}
The above equation actually enumerates and sums the computation of each neuron in every expert. Below, we omit the expert and neuron indices ($i$ and $j$), and instead use a unified index $i$ to enumerate all neurons. Additionally, let the row vectors of matrices $\mathbf{W}^\mathrm{up}$ and $\mathbf{W}^\mathrm{down}$ be denoted as $\mathbf{w}^\mathrm{up}$ and $\mathbf{w}^\mathrm{gate}$, and the column vectors of matrix $\mathbf{W}^\mathrm{down}$ also be denoted as $\mathbf{w}_i^\mathrm{down}$. The equation above can then be rewritten as:
\begin{align}
\mathbf{y} &= \sum_i^{N_E\times d_\mathrm{ff}}A_i\cdot \sigma\left(\mathbf{w}_i^{\mathrm{gate}}\mathbf{x}\right)\cdot\mathbf{w}_i^{\mathrm{up}}\mathbf{x}\cdot\mathbf{w}^{\mathrm{down}}.\notag
\end{align}
We separate the scalar terms in the above equation and denote them as $\phi_i = A_i\cdot \sigma\left(\mathbf{w}_i^{\mathrm{gate}}\mathbf{x}\right)\cdot\mathbf{w}_i^{\mathrm{up}}\mathbf{x}$. Hence, we have:
\begin{align}
\mathbf{y} &=\sum_i^{N_e}\phi_i\cdot\mathbf{w}_i^{\mathrm{down}}.\notag
\end{align}
This is the general form of mixed micro-experts as presented in Equation~\ref{eq:mome} of Section~\ref{subsec:me}.

\subsection{Proof Details}
\label{subsec:proof}

\paragraph{Lemma} Given the activation coefficient matrix $\mathbf{\Phi}$ and the matrix of basis vectors $\mathbf{W}$, let $S^C$ be the index set of removed micro-experts. Then the upper-bound of the decoding error $\epsilon$ on the calibration set is given by:
\begin{equation}
    \epsilon_{\mathrm{sup}} = \sum_{i\in S^C}\|\mathbf{\Phi}_{:,i}\|_2^2\|\mathbf{w}_i\|_2^2.\notag
\end{equation}
\paragraph{Proof} From Equation~\ref{eq:cssp}, we can see that the decoding error of the MoE layer is $\epsilon=\|\mathbf{\Phi W}-\mathbf{\Phi}_{:,S}\mathbf{W}_{S,:}\|_F^2=\|\mathbf{\Phi}_{:,S^C}\mathbf{W}_{S^C,:}\|_F^2$. Assuming the calibration dataset contains $n$ tokens, the Frobenius norm of the matrix can be equivalently expressed as a sum over rows:
\begin{equation}
    \epsilon = \|\mathbf{\Phi}_{:,S^C}\mathbf{W}_{S^C,:}\|_F^2=\sum_{j=1}^n \|\sum_{i\in S^C}\phi_{ji}\mathbf{w}_i\|_2^2.\notag
\end{equation}
By the Minkowski inequality, we have:
\begin{equation}
    \sum_{j=1}^n \|\sum_{i\in S^C}\phi_{ji}\mathbf{w}_i\|_2^2\leq\sum_{j=1}^n \left(\sum_{i\in S^C}|\phi_{ji}|\cdot\|\mathbf{w}_i\|_2\right)^2.\notag
\end{equation}
By the Cauchy-Schwarz inequality, we have:
\begin{equation}
    \sum_{j=1}^n \left(\sum_{i\in S^C}|\phi_{ji}|\cdot\|\mathbf{w}_i\|_2\right)^2\leq \sum_{j=1}^n\left(\sum_{i\in S^C}|\phi_{ji}|^2\right)\cdot\left(\sum_{i\in S^C}\|\mathbf{w}_i\|_2^2\right)=\left(\sum_{j=1}^n\sum_{i\in S^C}|\phi_{ji}|^2\right)\cdot\left(\sum_{i\in S^C}\|\mathbf{w}_i\|_2^2\right).\notag
\end{equation}
Note that
\begin{equation}
    \sum_{j=1}^n\sum_{i\in S^C}|\phi_{ji}|^2=\sum_{i\in S^C}\sum_{j=1}^n|\phi_{ji}|^2=\sum_{i\in S^C}\|\mathbf{\Phi}_{:,i}\|_2^2.\notag
\end{equation}
Hence, we can finally get:
\begin{equation}
    \epsilon=\|\mathbf{\Phi}_{:,S^C}\mathbf{W}_{S^C,:}\|_F^2\leq\sum_{i\in S^C}\|\mathbf{\Phi}_{:,i}\|_2^2\|\mathbf{w}_i\|_2^2=\epsilon_\mathrm{sup}.\notag
\end{equation}
The lemma is proved.

\begin{figure*}[t!]
\centering
\includegraphics[width=0.75\textwidth]{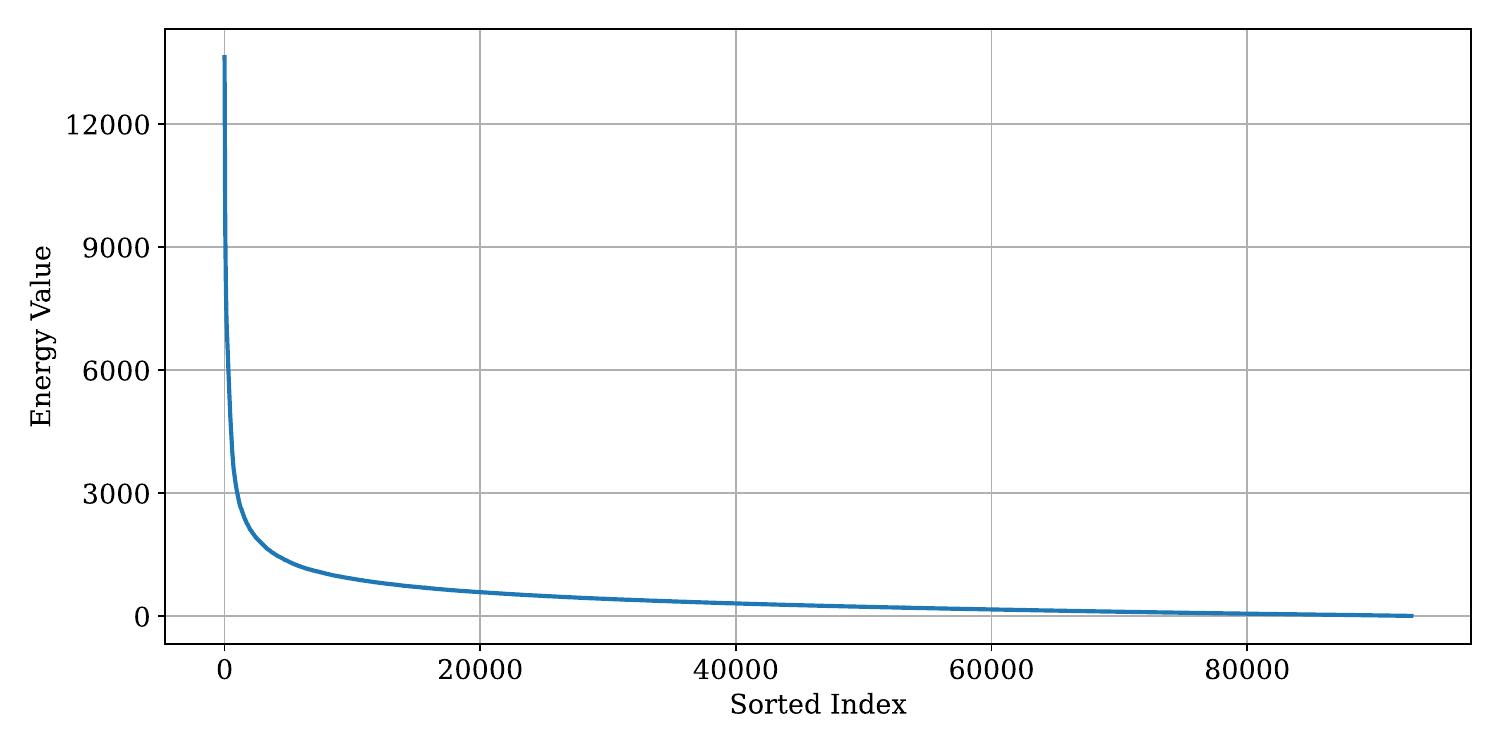}
\caption{Energy distribution of micro-experts in layer 12 of Deepseek-MoE-16B. For clarity, we reorder all micro-experts by the absolute value of their energy and remove the top-50 micro-experts with the extremely highest energy.}
\label{fig:ener_dist}
\end{figure*}

\paragraph{Theorem} Let $\hat{\mathbf{Y}}$ denote the decoding result using the Top-$|S|$ highest-energy micro-experts, and let $\mathbf{Y}^{*}$ denote the rank-$|S|$ SVD approximation of $\mathbf{Y}$. If $k=N_e-|S|$, the approximation error of $\hat{\mathbf{Y}}$ differs from the optimal SVD by only an $O(k)$-delta, i.e.,
\begin{equation}
    \|\mathbf{Y} - \hat{\mathbf{Y}}\|_F^2 \leq \|\mathbf{Y} - \mathbf{Y}^{*}\|_F^2 + \delta\left(O(k)\right).\notag
\end{equation}
\paragraph{Proof} In the \Camera algorithm, we define the ``inference-time energy'' of each micro-expert. In practice, we rank all micro-experts by their energy and prune the trailing ones (i.e., the last $k$ micro-experts). For convenience, we first sort the matrices $\mathbf{\Phi}$ and $\mathbf{W}$ based on the energy of their corresponding micro-experts. After sorting, we have $\mathcal{E}_1 \geq \mathcal{E}_2 \geq \cdots \geq \mathcal{E}_{|S|}=\varepsilon \geq \cdots \geq \mathcal{E}_{N_e}$, where $\varepsilon$ is the boundary energy. For each $i>|S|$, its energy is below this bound $\mathcal{E}_i=\|\mathbf{\Phi}_{:,i}\|_2^2\|\mathbf{w}_i\|_2^2\leq\varepsilon$. From the lemma above, we have:
\begin{equation}
    \|\mathbf{Y} - \hat{\mathbf{Y}}\|_F^2 \leq \sum_{i>|S|}\varepsilon.\notag
\end{equation}
By the Eckart-Young-Mirsky theorem, we know that the truncated SVD gives the matrix with the closest distance to $\mathbf{Y}$ among all matrices whose rank does not exceed $|S|$. The approximation error is:
\begin{equation}
    \|\mathbf{Y} - \mathbf{Y}^{*}\|_F^2 = \sum_{i>|S|}\sigma_i^2,\notag
\end{equation}
where $\sigma_i$ is the \textit{i}-th singular value of matrix $\mathbf{Y}$. We can then compare the difference between the two errors above:
\begin{equation}
    \|\mathbf{Y} - \hat{\mathbf{Y}}\|_F^2 - \|\mathbf{Y} - \mathbf{Y}^{*}\|_F^2 \leq \sum_{i>|S|}\varepsilon - \sum_{i>|S|}\sigma_i^2 \leq \sum_{i>|S|}\left(\varepsilon-\min_{i>|S|}\sigma_i^2\right)=k\left(\varepsilon-\min_{i>|S|}\sigma_i^2\right).\notag
\end{equation}
Let $\delta\left(O(k)\right)=k\left(\varepsilon-\min_{i>|S|}\sigma_i^2\right)$, we have:
\begin{equation}
    \|\mathbf{Y} - \hat{\mathbf{Y}}\|_F^2 \leq \|\mathbf{Y} - \mathbf{Y}^{*}\|_F^2 + \delta\left(O(k)\right).\notag
\end{equation}
The theorem is proved. This shows that the decoding error of the MoE layer after micro-expert pruning differs from the optimal rank‑$|S|$ approximation by at most a value $\delta\left(O(k)\right)$ proportional to $k$. The magnitude of $\delta$ depends on the energy distribution of the micro-experts: if their energies vary significantly and the boundary energy $\varepsilon$ is faint, then $\delta$ will also be small. In such cases, the energy-based pruning algorithm \CameraP can achieve a good and controllable approximation. Figure~\ref{fig:ener_dist} confirms our assumption that the energy distribution of micro-experts is highly non-uniform, which ensures the effectiveness of \CameraP.

\subsection{Supplementary Micro-Expert Distributions}
\label{subsec:supdis}

\begin{figure*}[t]
\centering
\begin{subfigure}[t]{1.00\textwidth}
    \centering
    \includegraphics[width=\textwidth]{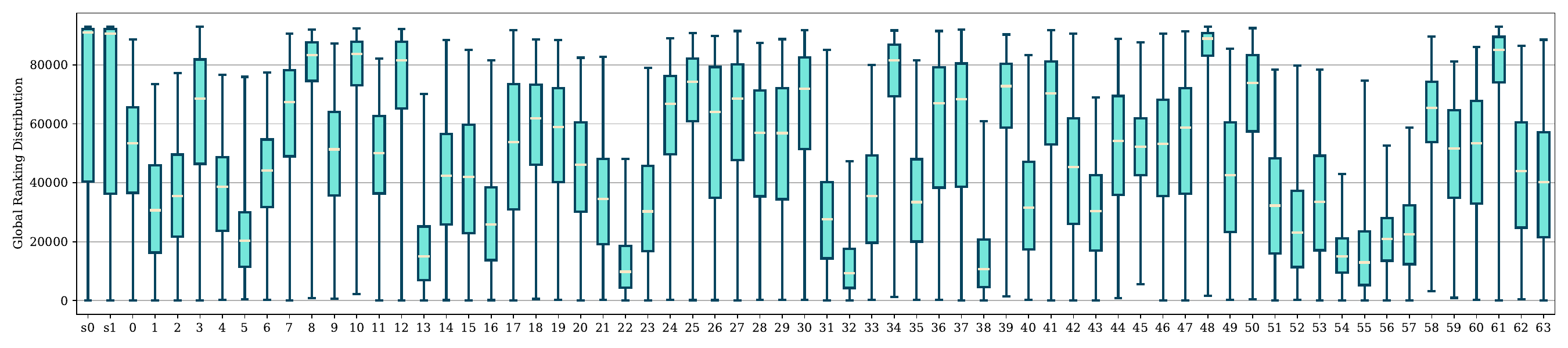}
    \caption{Layer 3, $\lambda=20\%$}
    \label{fig:d16b_3}
\end{subfigure}
\begin{subfigure}[t]{1.00\textwidth}
    \centering
    \includegraphics[width=\textwidth]{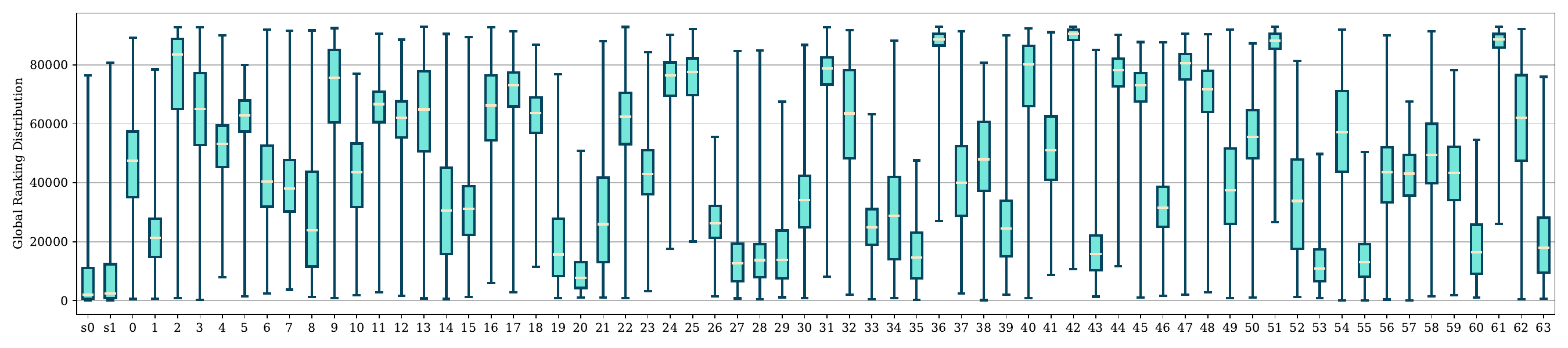}
    \caption{Layer 28, $\lambda=20\%$}
    \label{fig:d16b_28}
\end{subfigure}
\begin{subfigure}[t]{1.00\textwidth}
    \centering
    \includegraphics[width=\textwidth]{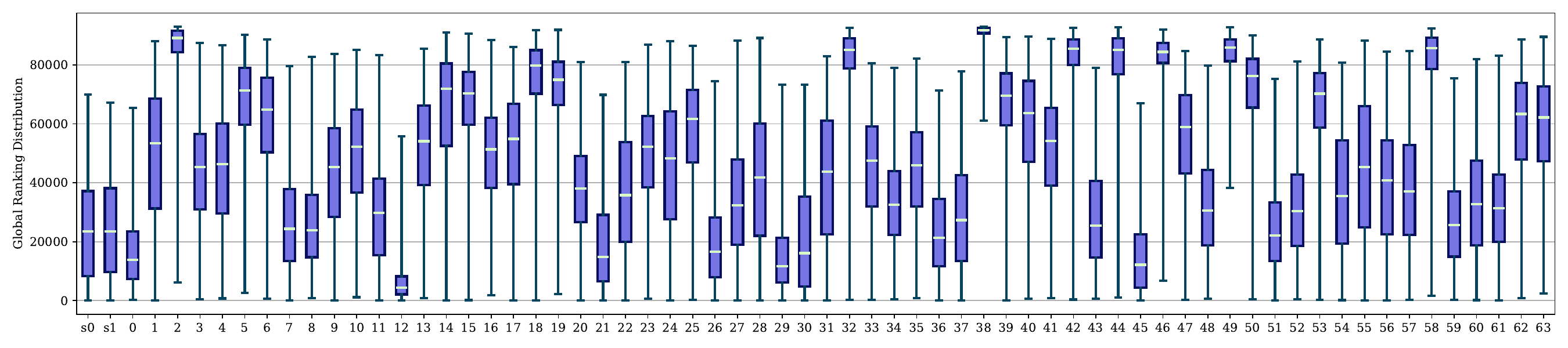}
    \caption{Layer 12, $\lambda=40\%$}
    \label{fig:d16b_12}
\end{subfigure}
\caption{Distributions of micro-experts within each expert based on global ranking from \Camera on Deepseek-MoE-16B. We list all 66 experts, where `S0/S1' denotes the shared experts, and the rest are non-shared common experts. For~\ref{fig:d16b_12}, we present the results of the pruned model at $\lambda = 40\%$, while the other subfigures correspond to $\lambda = 20\%$.}
\label{fig:d16b_others}
\end{figure*}

\paragraph{Layers of Deepseek-MoE-16B} In Section~\ref{subsec:medis} of the main text, we presented the micro-expert analysis of layer 12 in the Deepseek-MoE-16B model with a overall pruning ratio of $\lambda = 20\%$. Here, we further supplement the analysis with results from layers 3 and 28, as shown in Figures~\ref{fig:d16b_3} and~\ref{fig:d16b_28}. \textbf{First}, we observe that the importance of the first two shared experts, S0 and S1, is quite similar across layers, though not exactly the same. In contrast, the importance of the remaining experts varies significantly. \textbf{Additionally}, we also find that the shared experts tend to be less important in the lower layers (e.g., layers 2 and 3), as illustrated in Figure~\ref{fig:d16b_3}. \textbf{Finally}, we highlight a seemingly important property of the \Camera algorithm: the micro-expert analysis results—namely, the importance rankings—remain largely consistent across different values of $\lambda$. As shown in Figure~\ref{fig:d16b_12}, when $\lambda = 40\%$, the analysis results closely resemble those in Figure~\ref{fig:dist}.

\begin{figure*}[t]
\centering
\begin{subfigure}[t]{1.00\textwidth}
    \centering
    \includegraphics[width=\textwidth]{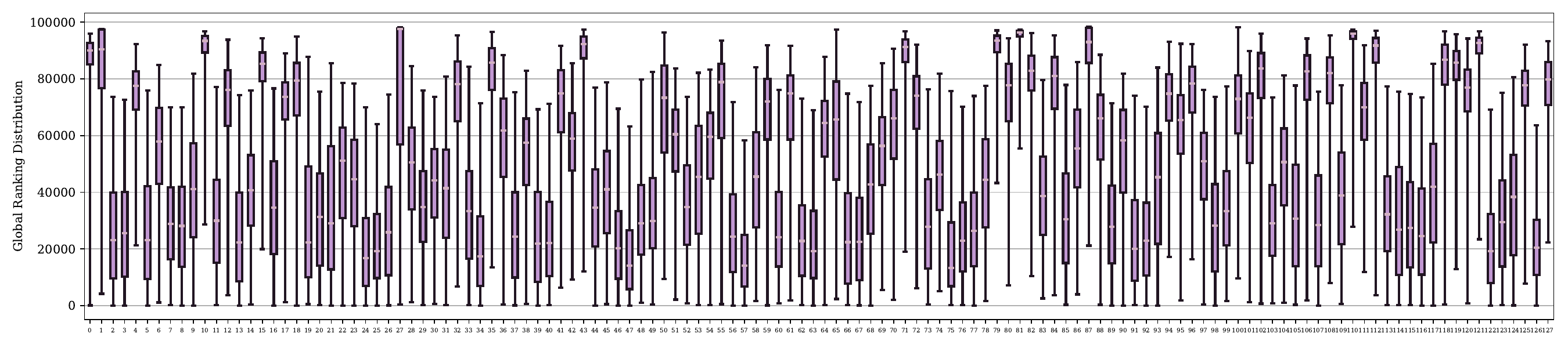}
    \caption{Layer 2, $\lambda=20\%$}
    \label{fig:q3_2}
\end{subfigure}
\begin{subfigure}[t]{1.00\textwidth}
    \centering
    \includegraphics[width=\textwidth]{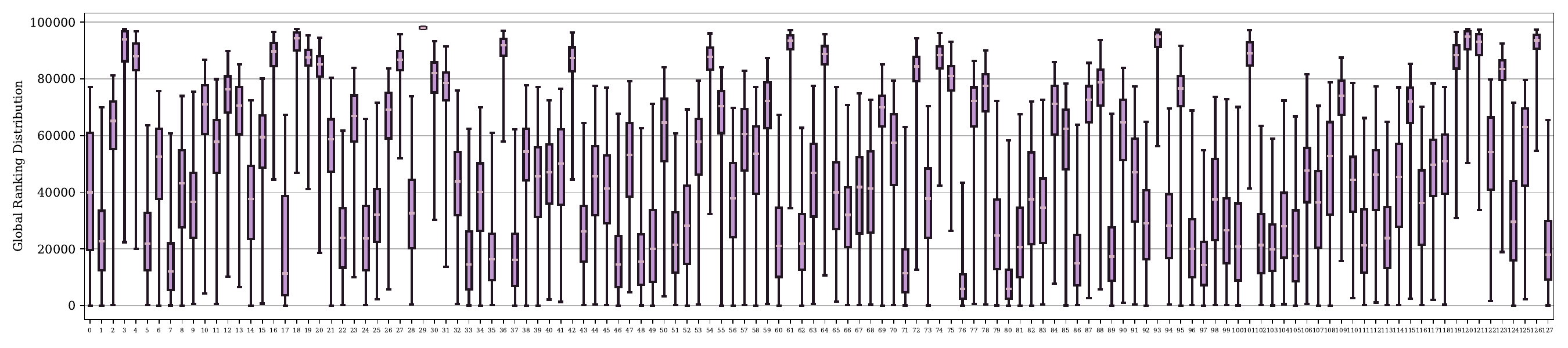}
    \caption{Layer 22, $\lambda=20\%$}
    \label{fig:q3_22}
\end{subfigure}
\begin{subfigure}[t]{1.00\textwidth}
    \centering
    \includegraphics[width=\textwidth]{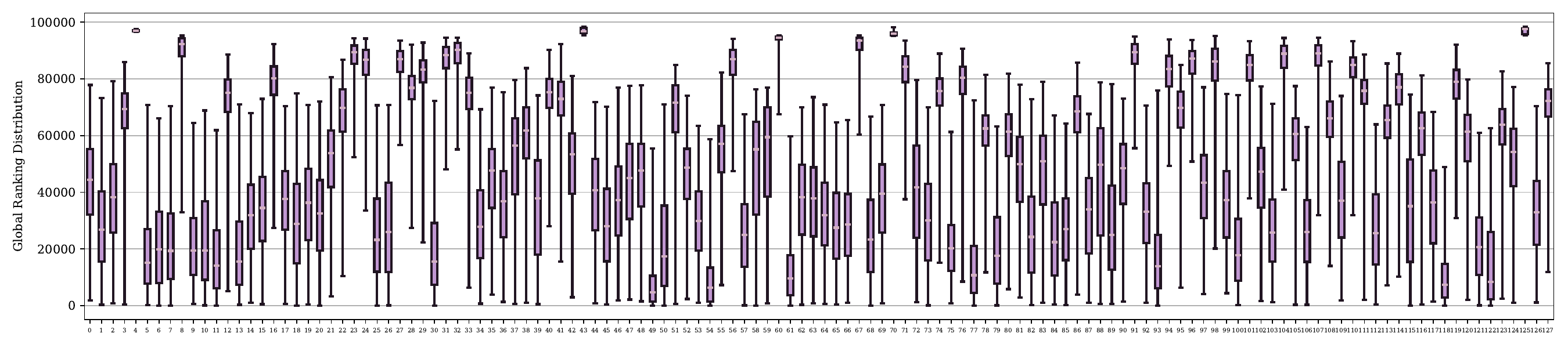}
    \caption{Layer 42, $\lambda=20\%$}
    \label{fig:q3_42}
\end{subfigure}
\caption{Distributions of micro-experts within each expert based on global ranking from \Camera on Qwen3-30B-A3B. We list all 128 non-shared common experts and the results are from the pruned model at $\lambda = 20\%$.}
\label{fig:qwen3}
\end{figure*}

\paragraph{Layers of Qwen3-30B-A3B} As shown in Figure~\ref{fig:qwen3}, the micro-expert distribution of the Qwen3-30B-A3B model is similar to that of Deepseek-MoE-16B, exhibiting large disparities in expert importance. Unlike Deepseek-MoE-16B, Qwen3-30B-A3B does not employ shared experts—all 128 experts are treated equally. Moreover, we observe that Qwen3-30B-A3B appears to have a higher proportion of unimportant experts. As shown in Figures~\ref{fig:q3_22} and~\ref{fig:q3_42}, more expert box plots are concentrated near the top. We speculate that this may result from insufficient weight utilization, suggesting potential for continued training and further knowledge injection.

\begin{table*}[t!]
\centering
\small
\begin{tabular}{c@{\hspace{0.20cm}}|@{\hspace{0.40cm}}c@{\hspace{0.40cm}}|c@{\hspace{0.6cm}}c|c@{\hspace{0.30cm}}c@{\hspace{0.30cm}}c@{\hspace{0.30cm}}c@{\hspace{0.30cm}}c@{\hspace{0.30cm}}c@{\hspace{0.30cm}}c@{\hspace{0.30cm}}c@{\hspace{0.30cm}}c@{\hspace{0.30cm}}c}
\toprule
$\lambda$ & Method & Wiki2 & C4 & BoolQ & OBQA & RTE & Wino. & Hella. & PIQA & MathQA & ARC-e & ARC-c & Avg. \\
\midrule
\multicolumn{14}{c}{\textbf{Phi3.5-MoE-42B}\quad(16 common experts, top-2 experts are activated)}\\
\cmidrule(lr){1-14}
0\% & Original & 3.99 & 7.66 & 88.38 & 51.00 & 77.26 & 76.01 & 79.73 & 77.53 & 38.06 & 65.70 & 53.84 & 67.50 \\
\cmidrule(lr){1-14}
\multirow{3}{*}[-0ex]{25\%} & NAEE & 4.55 & \textbf{8.41} & 86.91 & 48.60 & 76.53 & 76.32 & \textbf{80.68} & 73.61 & 33.77 & 54.34 & 46.59 & 64.15 \\
& $D^2$-MoE & 5.59 & 10.96 & 85.57 & 47.00 & 75.45 & 74.19 & 79.85 & 73.61 & 34.04 & 57.37 & 50.09 & 64.13 \\
& \CameraP & \textbf{4.33} & 8.43 & \textbf{87.77} & \textbf{50.60} & \textbf{76.53} & \textbf{76.69} & 79.40 & \textbf{75.90} & \textbf{36.15} & \textbf{61.87} & \textbf{51.62} & \textbf{66.28} \\
\midrule
\multicolumn{14}{c}{\textbf{Mixtral-8$\times$7B}\quad(8 common experts, top-2 experts are activated)}\\
\cmidrule(lr){1-14}
0\% & Original & 3.84 & 6.88 & 85.44 & 46.20 & 70.40 & 75.85 & 84.05 & 83.51 & 41.34 & 83.59 & 59.90 & 70.03 \\
\cmidrule(lr){1-14}
\multirow{3}{*}[-0ex]{25\%} & NAEE & 4.98 & \textbf{7.77} & 83.06 & 46.40 & 67.15 & \textbf{75.53} & \textbf{81.35} & \textbf{82.10} & \textbf{39.66} & \textbf{79.80} & 55.03 & \textbf{67.79} \\
& $D^2$-MoE & \textbf{4.90} & 8.80 & 82.20 & 45.60 & 66.85 & 74.51 & 79.93 & 81.28 & 36.55 & 77.82 & 52.65 & 66.38 \\
& \CameraP & 4.96 & 8.89 & \textbf{84.37} & \textbf{47.60} & \textbf{71.12} & 71.43 & 79.36 & 78.84 & 38.69 & 79.55 & \textbf{56.48} & 67.49 \\
\bottomrule
\end{tabular}
\caption{Supplementary pruning results of evaluation experiment on Phi3.5-MoE-42B and Mixtral-8$\times$7B. The best scores are in bold. We also test the performance of the original model (16-bit) as a reference.}
\label{tab:mixtral}
\end{table*}

\begin{table*}[t!]
\centering
\small
\begin{tabular}{@{\hspace{0.40cm}}c@{\hspace{0.40cm}}|c@{\hspace{0.50cm}}c@{\hspace{0.50cm}}c@{\hspace{0.50cm}}c@{\hspace{0.50cm}}c}
\toprule
Model & Mixtral-8$\times$7B & Phi3.5-MoE-42B & Deepseek-MoE-16B & Qwen3-30B-A3B & Deepseek-V2 \\
\midrule
Expert Setting & 8 & 16 & 2+64 & 128 & 2+160 \\
Act. Setting & top2 & top2 & 2+top4 & top8 & 2+top6 \\
$r_\mathrm{act}$ & 25.00\% & 12.50\% & 6.25\% & 6.25\% & 3.75\% \\
$p_\mathrm{lossless}$ & 53.57\% & 55.00\% & 30.62\% & 9.27\% & 17.24\% \\
\bottomrule
\end{tabular}
\caption{Comparison of some mainstream MoE models. $r_\mathrm{act}$ denotes the activation ratio of expert weight within MoE layers. $p_\mathrm{lossless}$ indicates the probability that the MoE remains losslessly activated after reducing 25\% of experts using NAEE.}
\label{tab:models}
\end{table*}

\subsection{Supplementary Pruning Experiments}
\label{subsec:supexp}

In our primary experiments, we evaluate the effectiveness of the \CameraP algorithm on models featuring a larger number of smaller experts. This design choice is strongly supported by recent developments in the open-source community. Over the past two years, leading organizations and research groups—including Alibaba (Qwen series) and Deepseek—have consistently advanced the state-of-the-art by adopting MoE architectures that mix more lightweight experts. Notable examples include Qwen1.5-MoE-A2.7B, Qwen2-57B-A14B, Qwen3-30B-A3B, and the Deepseek-MoE family. In contrast, earlier MoE models often employ fewer but heavier experts, leading to higher activation ratios and lower computational efficiency, as shown in Table~\ref{tab:models}. To further assess the performance of \CameraP under such settings ($N_E \leq 16$), we conduct additional experiments on Phi3.5-MoE-42B and Mixtral-8$\times$7B, as shown in Table~\ref{tab:mixtral}.

Our observation is that, in terms of perplexity and accuracy, the performance advantage of \CameraP appears to be less pronounced on these two models compared to other MoEs with more and smaller experts. We hypothesize that one possible reason is that these models, after being pruned by NAEE, exhibit a higher probability of lossless activation. If a model pruned by NAEE contains $0.75N_E$ experts and activates $N_A$ experts per forward pass, the lossless activation probability is
\begin{equation}
    p_\mathrm{lossless}=\frac{\binom{0.75N_E}{N_A} }{\binom{N_E}{N_A}}.\notag
\end{equation}
Table~\ref{tab:models} shows that NAEE achieves a clearly higher lossless activation probability $p_\mathrm{lossless}$ on the two models above, which may explain its better performance. Nevertheless, we would like to emphasize that models with lower activation ratios and a larger number of smaller experts represent the mainstream and future direction of MoE advancement.

\subsection{Mixed-precision Settings in \CameraQ}
\label{subsec:mixsetting}

In the experiments in Table~\ref{tab:cameraq} of the main text, MC assigns different precisions across experts, whereas \CameraQ and \CameraQd apply varying precisions across micro-experts within each expert. Consequently, the weight matrix of each expert consists of multiple precision levels. We implement \CameraQ and its variants using a three-level mixed-precision scheme. To achieve an average bit-width of $b$, we define three precision levels: $[b+1, b, b-1]$ bits, with their respective proportions in the total weights set to $[r, 1-2r, r]$. For example, if the desired average precision is 2 bits with $r=0.1$, then the bit-widths $[3, 2, 1]$ are assigned in proportions $[0.1, 0.8, 0.1]$, yielding an average of $3\times0.1+2\times0.8+1\times0.1=2.0$ bits. Since we perform group-wise quantization over every 128 vectors, each group stores an additional zero-point and scale factor, totaling $2\times16=32$ bits (following the GPTQ algorithm~\citep{gptq2022}). This adds an overhead of $32/128=0.25$ bit per weight, resulting in a total average cost of 2.25 bits per weight.

\begin{figure*}[t]
\centering
\includegraphics[width=0.90\textwidth]{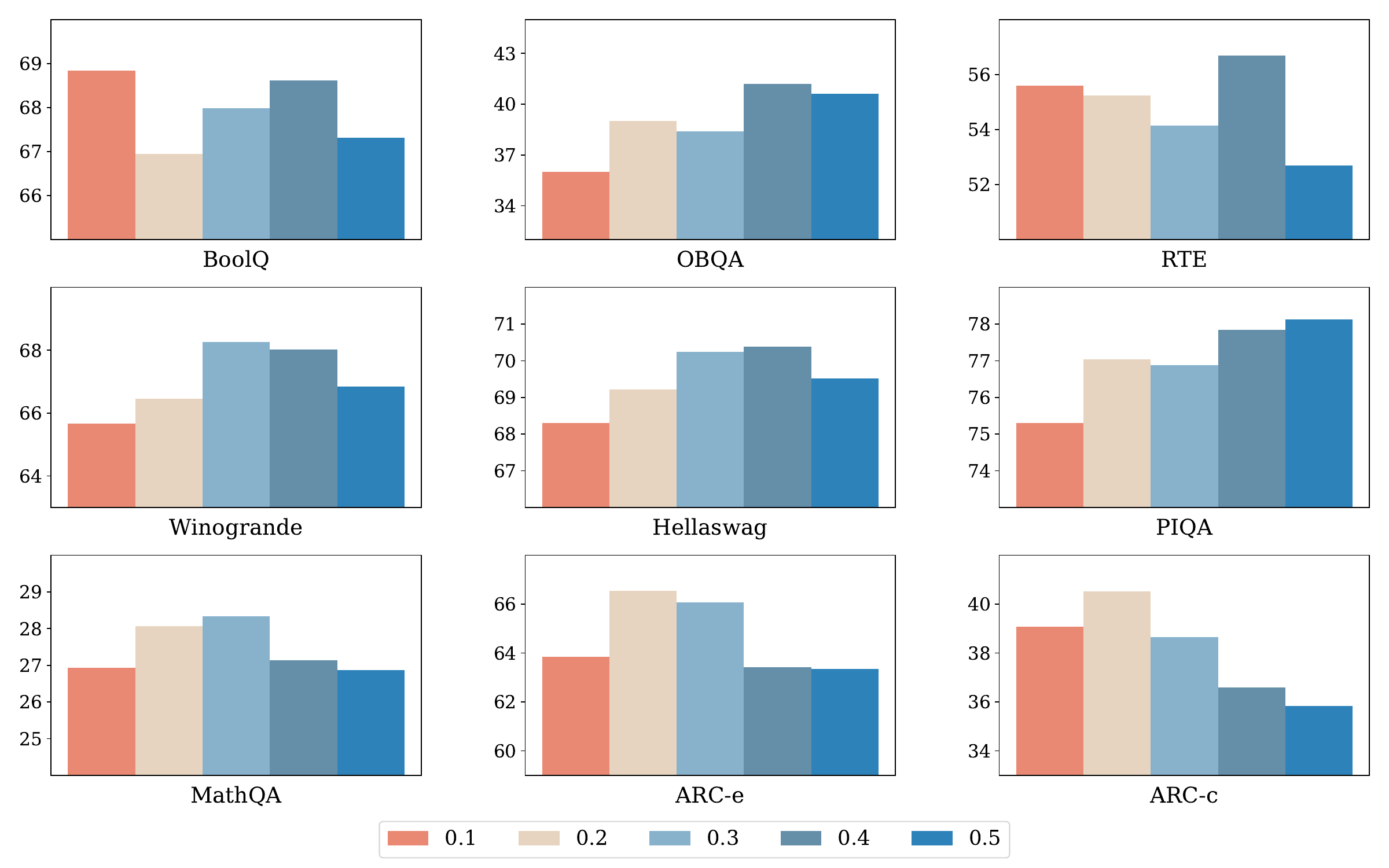}
\caption{Downstream task performance of \CameraQ on Deepseek-MoE-16B under different $r$ from 0.1 to 0.5. At $r=0.5$, the three-level mixed-precision reduces to two-level.}
\label{fig:quant_tasks}
\end{figure*}

The mixed-precision configuration described above involves a hyperparameter $r$, which determines the proportion of high- and low-precision assignments. This parameter not only influences the final task performance but also provides insight into how mixed-precision impacts quantization effectiveness. To explore this, we evaluate the quantization performance of \CameraQ under various $r$. Figure~\ref{fig:quant_tasks} shows the downstream task scores across five different $r$ settings. Notably, neither increasing nor decreasing $r$ monotonically improves performance. In most cases, the best results are achieved when $r$ is set to 0.2 or 0.3. This highlights the effectiveness of a micro-expert-oriented mixed-precision idea, where higher precision is allocated to a small subset of important micro-experts. In the experiments reported in Table~\ref{tab:cameraq}, both \CameraQ and \CameraQd adopt $\mathbf{r=0.2}$, using a group quantization size of \textbf{128}.

\begin{table*}[t!]
\centering
\small
\begin{tabular}{@{\hspace{0.40cm}}c@{\hspace{0.40cm}}c@{\hspace{0.40cm}}c@{\hspace{0.40cm}}c@{\hspace{0.40cm}}c@{\hspace{0.40cm}}c@{\hspace{0.40cm}}c@{\hspace{0.40cm}}c@{\hspace{0.40cm}}c@{\hspace{0.40cm}}c@{\hspace{0.40cm}}c@{\hspace{0.40cm}}c@{\hspace{0.40cm}}c@{\hspace{0.40cm}}c}
\toprule
Model & \multicolumn{6}{c}{Mixtral-8$\times$7B} & \multicolumn{6}{c}{Deepseek-MoE-16B} \\
\cmidrule(lr){1-1} \cmidrule(r){2-7} \cmidrule(r){8-13}
Layer & \multicolumn{2}{c}{Low} & \multicolumn{2}{c}{Medium} & \multicolumn{2}{c}{High} & \multicolumn{2}{c}{Low} & \multicolumn{2}{c}{Medium} & \multicolumn{2}{c}{High} \\
\cmidrule(lr){1-1} \cmidrule(r){2-3} \cmidrule(r){4-5} \cmidrule(r){6-7} \cmidrule(r){8-9} \cmidrule(r){10-11} \cmidrule(r){12-13}
Metric & L2 & Cosine & L2 & Cosine & L2 & Cosine & L2 & Cosine & L2 & Cosine & L2 & Cosine \\
\cmidrule(lr){1-1} \cmidrule(r){2-3} \cmidrule(r){4-5} \cmidrule(r){6-7} \cmidrule(r){8-9} \cmidrule(r){10-11} \cmidrule(r){12-13}
NAEE & 26.7 & 0.935 & 983.6 & 0.765 & 15063.1 & 0.859 & 53.2 & 0.958 & 600.5 & 0.819 & 741.0 & 0.844 \\
$D^2$-MoE & 33.9 & 0.940 & 1350.1 & 0.728 & 22907.7 & 0.811 & 42.4 & 0.963 & 561.3 & 0.885 & 982.8 & 0.821 \\
\CameraP & 21.0 & 0.988 & 1486.5 & 0.718 & 10575.9 & 0.864 & 31.0 & 0.999 & 316.8 & 0.918 & 585.2 & 0.875 \\
\bottomrule
\end{tabular}
\caption{Comparison of intermediate hidden states after pruning from different algorithms. The experimental setup follows previous sections. For Mixtral-8$\times$7B, we prune 25\% of the experts and report results for layers 1, 16, and 31; for Deepseek-MoE-16B, we prune 20\% and report layers 2, 12, and 22. ``L2'' denotes the L2 norm of the difference between the pruned hidden states and those of the original model, while ``Cosine'' refers to the cosine similarity between them.}
\label{tab:errors}
\end{table*}

\subsection{Approximation Errors}
\label{subsec:apperr}

We conduct a quantitative analysis of the pruning performance of several algorithms. As shown in Table~\ref{tab:errors}, we select three layers from each of the two models and compare the approximation error of the pruned MoE layer outputs to those of the original model across three pruning methods. Two metrics are used: L2 distance and cosine similarity. \CameraP demonstrates strong approximation to the original model in both shallow and deep layers, though it may not achieve optimal results in some intermediate layers. Moreover, we observe that the numerical analysis does not always align with downstream task performance. In Appendix~\ref{subsec:supexp}, we provide a possible explanation: a higher proportion of lossless tokens may improve task scores, even when the numerical approximation is not the most accurate.

\subsection{Details on Baselines}
\label{subsec:details}

We provide the essential supplemental details of the baselines in this work to facilitate the reproduction of the results:
\begin{itemize}
    \item NAEE: We adopt a layer-wise pruning strategy without dynamic layer skipping. The calibration dataset and data size are kept consistent with \CameraP. For the genetic algorithm variant of NAEE, we set the initial population size to 100, preserve the top 20\% as elites, use a mutation probability of 0.5, and run for 100 generations. The observed loss reduction trend indicates a favorable approximation performance.
    \item $D^2$-MoE: We use the official open-source implementation. The calibration dataset is of the same source as that used in \CameraP, with the data size following the open-source recommendation (approximately 5$\sim$10x larger than that of \CameraP). The shared base weight are selected based on the \textit{Fisher} criterion, with \texttt{pp\_ratio} set to 0.1 to preserve model capacity as much as possible. The \texttt{delta\_ratio} is determined using the search code from the author, and \texttt{share\_ratio} is set to 1.
    \item MC: Similar to \CameraQ, MC uses the C4 dataset for calibration to compute expert activation and pre-quantization statistics. Only the MoE layers are quantized, while the attention layers remain in full-precision. The final quantization is based on GPTQ, with a group size of 128 for group-wise quantization.
\end{itemize}

\end{document}